\documentclass{article}

\usepackage{microtype}
\usepackage{graphicx}
\usepackage{subfigure}
\usepackage{booktabs} 
\usepackage{array}
\usepackage{amsfonts,amsmath}

\usepackage{hyperref, url}

\usepackage[accepted]{icml2019}

\icmltitlerunning{Assessing Generalization in Deep Reinforcement Learning}

\usepackage[inline]{enumitem}

\newcommand\mypara[1]{\vspace{1mm}\noindent\textbf{#1}}

\newcommand\timess{\mathbin{\!\times\!}}
\newif\ifisaccepted

\newcommand{\e}{\mathbb{E}}

\newcommand{\reals}{\mathbb{R}}
\newcommand{\aA}{\mathcal{A}}

\newcommand{\sS}{\mathcal{S}}

\newcommand{\bs}{\mathbf{s}}
\newcommand{\ba}{\mathbf{a}}
\newcommand{\br}{\mathbf{r}}

\begin{document}

\twocolumn[
\icmltitle{Assessing Generalization in Deep Reinforcement Learning}

\icmlsetsymbol{equal}{*}

\begin{icmlauthorlist}
\icmlauthor{Charles Packer}{equal,berk}
\icmlauthor{Katelyn Gao}{equal,intel}
\icmlauthor{Jernej Kos}{nus}
\icmlauthor{Philipp Kr\"ahenb\"uhl}{texas}
\icmlauthor{Vladlen Koltun}{intel}
\icmlauthor{Dawn Song}{berk}
\end{icmlauthorlist}

\icmlaffiliation{berk}{University of California Berkeley, Berkeley, California, USA}
\icmlaffiliation{intel}{Intel Labs, Santa Clara, California, USA}
\icmlaffiliation{nus}{National University of Singapore, Singapore}
\icmlaffiliation{texas}{University of Texas Austin, Austin, Texas, USA}

\icmlcorrespondingauthor{Charles Packer}{cpacker@berkeley.edu}
\icmlcorrespondingauthor{Katelyn Gao}{katelyn.gao@intel.com}

\icmlkeywords{generalization, reinforcement learning, benchmark}

\vskip 0.3in
]

\printAffiliationsAndNotice{\icmlEqualContribution}

\begin{abstract}
Deep reinforcement learning (RL) has achieved breakthrough results on many tasks, but agents often fail to generalize beyond the environment they were trained in. As a result, deep RL algorithms that promote generalization are receiving increasing attention. However, works in this area use a wide variety of tasks and experimental setups for evaluation. The literature lacks a controlled assessment of the merits of different generalization schemes. Our aim is to catalyze community-wide progress on generalization in deep RL. To this end, we present a benchmark and experimental protocol, and conduct a systematic empirical study. Our framework contains a diverse set of environments, our methodology covers both in-distribution and out-of-distribution generalization, and our evaluation includes deep RL algorithms that specifically tackle generalization. Our key finding is that ``vanilla'' deep RL algorithms generalize better than specialized schemes that were proposed specifically to tackle generalization.
\end{abstract}

\isacceptedtrue

\section{Introduction}
\label{sec:intro}

Deep reinforcement learning (RL) has emerged as an important family of techniques that may support the development of intelligent systems that learn to accomplish goals in a variety of complex real-world environments~\citep{Mnih2015dqn, Arulkumaran2017}.
A desirable characteristic of such intelligent systems is the ability to function in diverse environments, including ones that have never been encountered before.
Yet deep RL algorithms are commonly trained and evaluated on a fixed environment.
That is, the algorithms are evaluated in terms of their ability to optimize a policy in a complex environment, rather than their ability to learn a representation that generalizes to previously unseen circumstances.
The dangers of overfitting to a specific environment have been noted in the literature~\citep{Whiteson2011} and the sensitivity of deep RL to even subtle changes in the environment has been noted in recent work~\citep{rajeswaran2017towards, Henderson2017, zhang2018dissection}.

Generalization is often regarded as an essential characteristic of advanced intelligent systems and a central issue in AI research~\citep{Lake2017, Marcus2018, Dietterich2017}.
It includes both interpolation to environments similar to those seen during training and extrapolation outside the training distribution.
The latter is particularly challenging but is crucial to the deployment of systems in the real world as they must be able to handle unforseen situations.

Generalization in deep RL has been recognized as an important problem and is under active investigation~\citep{learningtorl, rlsquared, epopt, pintorobust, maml, Kansky2017, yu2017, sung2017learning, leike2017gridwords, continuousadaptation, clavera2018, saemundsson2018}.
However, each work uses a different set of environments, variations, and experimental protocols.
For example, \citet{Kansky2017} propose a graphical model architecture, evaluating on variations of the Atari game Breakout where the positions of the paddle, balls, and obstacles (if any) change.
\citet{epopt} propose training on a distribution of environments in a risk-averse manner and evaluate on two robot locomotion environments where variations in the robot body are considered.
\citet{rlsquared} aim to learn a policy that automatically adapts to the environment dynamics and evaluate on bandits, tabular Markov decision processes, and maze navigation environments where the goal position changes.
\citet{maml} train a policy that at test time can be quickly updated to the test environment and evaluate on navigation and locomotion.

What appears to be missing is a systematic empirical study of generalization in deep RL with a clearly defined set of environments, metrics, and baselines.
In fact, there is no common testbed for evaluating generalization in deep RL. Such testbeds have proven to be effective catalysts of concerted community-wide progress in other fields~\citep{Donoho2015}.
Only by conducting systematic evaluations on reliable testbeds can we fairly compare and contrast the merits of different algorithms and accurately assess progress.

Our contribution is an empirical evaluation of the generalization performance of deep RL algorithms. In doing so, we also establish a reproducible framework for investigating generalization in deep RL with the hope that it will catalyze progress on this problem.
Like \citet{Kansky2017}, \citet{epopt}, and others, we focus on generalization to environmental changes that affect the system dynamics instead of the goal or rewards.
We select a diverse set of environments that have been widely used in previous work on generalization in deep RL, comprising classic control problems and MuJoCo locomotion tasks, built on top of OpenAI Gym for ease of adoption. 
The environmental changes that affect the system dynamics are implemented by selecting degrees of freedom (parameters) along which the environment specifications can be varied.
Significantly, we test generalization in two regimes: interpolation and extrapolation.
Interpolation implies that agents should perform well in test environments where parameters are similar to those seen during training.
Extrapolation requires agents to perform well in test environments where parameters are different from those seen during training.

To provide the community with a set of clear baselines, we first evaluate two popular state-of-the-art deep RL algorithms under different combinations of training and testing regimes.
We choose one algorithm from each of the two major families: A2C from the actor-critic family and PPO from the policy gradient family.
Under the same experimental protocol, we also evaluate two recently-proposed schemes for tackling generalization in deep RL: EPOpt, which learns a policy that is robust to environment changes by maximizing expected reward over the most difficult of a distribution of environment parameters, and RL$^2$, which learns a policy that can adapt to the environment at hand by learning environmental characteristics from the trajectory it sees on-the-fly.
Because each scheme is constructed based on existing deep RL algorithms, our evaluation is of four algorithms: EPOpt-A2C, EPOpt-PPO, RL$^2$-A2C, and RL$^2$-PPO\@. 
We analyze the results, devising simple metrics for generalization in terms of both interpolation and extrapolation and drawing conclusions that can guide future work on generalization in deep RL\@.
The experiments confirm that extrapolation is more difficult than interpolation.

However, surprisingly, the vanilla deep RL algorithms, A2C and PPO, generalized better than their EPOpt and RL$^2$ variants; they were able to interpolate fairly successfully.
That is, simply training on a set of environments with variations can yield agents that can generalize to environments with similar variations.
EPOpt was able to improve generalization (both interpolation and extrapolation) but only when combined with PPO on environments with continuous action spaces and only one of the two policy/value function architectures we consider.
RL$^2$-A2C and RL$^2$-PPO proved to be difficult to train and were unable to reach the level of performance of the other algorithms given the same amount of training resources.
We discuss lines of inquiry that are opened by these observations.

\section{Related work}
\label{sec:related}

\mypara{Generalization in RL}. There are two main approaches to generalization in RL: learning policies that are robust to environment variations and learning policies that adapt to such variations.
A popular approach to learn a robust policy is to maximize a risk-sensitive objective, such as the conditional value at risk~\citep{cvar}, over a distribution of environments.
From a control theory perspective, \citet{morimoto2001} maximize the minimum reward over possible disturbances to the system model, proposing robust versions of the actor-critic and value gradient methods.
This maximin objective is utilized by others in the context where environment changes are modeled by uncertainties in the transition probability distribution function of a Markov decision process.
\citet{nilim} assume that the set of possible transition probability distribution functions are known, while \citet{limxumannor} and \citet{roymodelfree} estimate it using sampled trajectories from the distribution of environments of interest.
A recent representative of this approach applied to deep RL is the EPOpt algorithm~\citep{epopt}, which maximizes the conditional value at risk, i.e. expected reward over the subset of environments with lowest expected reward.
EPOpt has the advantage that it can be used in conjunction with any RL algorithm.
Adversarial training has also been proposed to learn a robust policy; for MuJoCo locomotion tasks, \citet{pintorobust} train an adversary that tries to destabilize the agent while it trains.

A robust policy may sacrifice performance on many environment variants in order to not fail on a few.
Thus, an alternative, recently popular approach to generalization in RL is to learn an agent that can adapt to the environment at hand \citep{yu2017}.
To do so, a number of algorithms learn an embedding for each environment variant using trajectories sampled from that environment, which is utilized by the agent.
Then, at test time, the current trajectory can be used to compute an embedding for the current environment, enabling automatic adaptation of the agent.
\citet{rlsquared}, \citet{learningtorl}, \citet{sung2017learning}, and \citet{mishra2017meta}, which differ mainly in the way embeddings are computed, consider model-free RL by letting the embedding be input into a policy and/or value function.
\citet{clavera2018} consider model-based RL, in which the embedding is input into a dynamics model and actions are selected using model predictive control.
Under a similar setup, \citet{saemundsson2018} utilize probabilistic dynamics models and inference.

The above approaches do not require updating the learned policy or model at test time, but there has also been work on generalization in RL that utilize such updates, primarily under the umbrellas of transfer learning, multi-task learning, and meta-learning.
\citet{TaylorStone2009} survey transfer learning in RL where a fixed test environment is considered, with \citet{Rusu2016} being an example of recent work on that problem using deep networks.
\citet{multitasksurvey} provides a survey of multi-task learning in general, which, different from our problem of interest, considers a fixed finite population of tasks.
\citet{maml} present a meta-learning formulation of generalization in RL, training a policy that can be updated with good data efficiency for each test environment; \citet{continuousadaptation} extend it for continuous adaptation in non-stationary environments.

\mypara{Empirical methodology in deep RL}. Shared open-source software infrastructure, which enables reproducible experiments, has been crucial to the success of deep RL\@.
The deep RL research community uses simulation frameworks, including OpenAI Gym~\citep{gym} (which we build upon), the Arcade Learning Environment~\citep{bellemareatari, Machado2017}, DeepMind Lab~\citep{deepmindlab}, and VizDoom~\citep{vizdoom}.
The MuJoCo physics simulator~\citep{mujoco} has been influential in standardizing a number of continuous control tasks.
\citet{leike2017gridwords} introduce a set of two-dimensional grid environments for RL, each designed to test specific safety properties of a trained agent, including in response to shifts in the distribution of test environments.
Recently OpenAI released benchmarks for generalization in RL based on playing new levels of video games, both allowing fine-tuning at test time~\citep{openairetro} and not~\citep{coinrun}.
Both \citet{coinrun} and \citet{justesen2018} (who also consider video games) use principled procedural generation of training and test levels based on difficulty.
In contrast to the above works, we focus on control tasks with no visual input.

Our work also follows in the footsteps of a number of empirical studies of reinforcement learning algorithms, which have primarily focused on the case where the agent is trained and tested on a fixed environment.
\citet{Henderson2017} investigate reproducibility in deep RL and conclude that care must be taken not to overfit during training. On four MuJoCo tasks, the results of state-of-the-art algorithms may be quite sensitive to hyperparameter settings, initializations, random seeds, and other implementation details.
The problem of overfitting in RL was recognized earlier by \citet{Whiteson2011}, who propose an evaluation methodology based on training and testing on multiple environments sampled from some distribution and experiment with three classic control environments and a form of tabular Q-learning.
\citet{rllab} present a benchmark suite of continuous control tasks and conduct a systematic evaluation of reinforcement learning algorithms on those tasks; they consider interpolation performance on a subset of their tasks.
\citet{leike2017gridwords} test generalization to unseen environment configurations at test time (referred to as `distributional shift') by varying the position of obstacles in a gridworld environment.
In contrast to these works, we consider a greater variety of tasks, extrapolation as well as interpolation, and algorithms for generalization in deep RL.

\section{Notation}
\label{sec:notation}

In RL, environments are formulated in terms of Markov Decision Processes (MDPs) \citep{suttonbarto}.
An MDP $M$ is defined by the tuple $\displaystyle (\sS, \aA, p, r, \gamma, \rho_0, T)$ where $\sS$ is the set of possible states, $\aA$ is the set of actions, $p: \sS \timess \aA \timess \sS \rightarrow \reals_{\geq 0}$ is the transition probability distribution function, $r: \sS \timess \aA \rightarrow \reals$ is the reward function, $\gamma$ is the discount factor, $\rho_0: \sS \rightarrow \reals_{\geq 0}$ is the initial state distribution at the beginning of each episode, and $T$ is the time horizon per episode.

Let $\bs_t$ and $\ba_t$ be the state and action taken at time $t$.
At the beginning of each episode, $\bs_0 \sim \rho_0(\cdot)$.
Under a policy $\pi$ stochastically mapping a sequence of states to actions, ${\ba_t \sim \pi(\ba_t \mid \bs_t, \cdots, \bs_0)}$ and $\bs_{t+1} \sim p(\bs_{t+1} \mid \ba_t)$, giving a trajectory $\{\bs_t,\ba_t,r(\bs_t,\ba_t)\}$, ${t=0,1,\cdots}$.
RL algorithms, taking the MDP as fixed, learn $\pi$ to maximize the expected reward (per episode) $J_M(\pi) = \e^\pi\left[\sum_{t=0}^T \gamma^t \br_t\right]$, where $\br_t=r(\bs_t, \ba_t)$.
They often utilize the concepts of a value function $v_M^\pi(\bs)$, the expected reward conditional on $\bs_0=\bs$ and a state-action value function $Q_M^\pi(\bs,\ba)$, the expected reward conditional on $\bs_0=\bs$ and $\ba_0=\ba$.

Algorithms that are designed to build RL agents that generalize often assume that there is a distribution of environments $q(M)$.
Then, they aim to learn a policy that maximizes the expected reward over the distribution, $\e_{M \sim q}^\pi\left[J_M(\pi)\right]$.

\section{Algorithms}
\label{sec:algs}

We first evaluate two popular state-of-the-art vanilla deep RL algorithms from different families, A2C~\citep{a3c} from the actor-critic family of algorithms and PPO~\citep{pporoboschool} from the policy gradient family.
\footnote{Preliminary experiments on other deep RL algorithms including A3C, TRPO, and ACKTR gave qualitatively similar results.}
These algorithms are oblivious to changes in the environment. 
Second, we consider two recently-proposed algorithms designed to train agents that generalize to environment variations.
We select one each from the two main types of approaches discussed in Section~\ref{sec:related}: EPOpt~\citep{epopt} from the robust category of approaches and RL$^2$~\citep{rlsquared} from the adaptive category.
Both methods are built on top of vanilla deep RL algorithms, so for completeness we evaluate a Cartesian product of the algorithms for generalization and the vanilla algorithms: EPOpt-A2C, EPOpt-PPO, RL$^2$-A2C, and RL$^2$-PPO\@.
Next we briefly summarize A2C, PPO, EPOpt, and RL$^2$, using the notation in Section~\ref{sec:notation}.

\mypara{Advantage Actor-Critic (A2C)}.
A2C involves the interplay of two optimizations; a critic learns a parametric value function, while an actor utilizes that value function to learn a parametric policy that maximizes expected reward.
At each iteration, trajectories are generated using the current policy, with the environment and hidden states of the value function and policy reset at the end of each episode.
Then, the policy and value function parameters are updated using RMSProp~\citep{rmsprop}, with an entropy term added to the policy objective function in order to encourage exploration.
We use an implementation from OpenAI Baselines~\citep{baselines}.

\mypara{Proximal Policy Optimization (PPO)}.
PPO aims to learn a sequence of monotonically improving parametric policies by maximizing a surrogate for the expected reward via gradient ascent, cautiously bounding the improvement at each iteration.
At iteration $i$, trajectories are generated using the current policy $\pi_{\theta_i}$, with the environment and hidden states of the policy reset at the end of each episode.
The following objective is then maximized with respect to $\theta$ using Adam~\citep{Kingma2015adam}:
\begin{equation*}
\resizebox{1\columnwidth}{!}{
\begin{math}
\e_{\bs \sim \rho_{\theta_i}, \ba \sim \pi_{\theta_i}}\min\left[\ell_{\theta}(\ba,\bs)A_{\pi_{\theta_i}}(\bs,\ba),m_{\theta}(\ba,\bs)A_{\pi_{\theta_i}}(\bs,\ba)\right]
\end{math}}
\end{equation*}
where $\rho_{\theta_i}$ are the expected visitation frequencies under $\pi_{\theta_i}$, $\ell_{\theta}(\ba,\bs)=\pi_{\theta}(\ba \mid \bs)/\pi_{\theta_i}(\ba \mid \bs)$, $m_{\theta}$ equals $\ell_{\theta}(\ba,\bs)$ clipped to the interval $[1-\delta,1+\delta]$ with $\delta \in (0,1)$, and $A_{\pi_{\theta_i}}(\bs,\ba)=Q_M^{\pi_{\theta_i}}(\bs,\ba)-v_M^{\pi_{\theta_i}}(\bs)$.
We use an implementation from OpenAI Baselines, PPO2.

\mypara{Ensemble Policy Optimization (EPOpt)}.
In order to obtain a policy that is robust to possibly out-of-distribution environments, EPOpt maximizes the expected reward over the $\epsilon \in (0,1]$ fraction of environments with worst expected reward:
\[
\e_{M \sim q}^\pi\left[J_M(\pi) \leq y \right] \text{\quad where \quad} P_{M \sim q} (J_M(\pi) \leq y) = \epsilon.
\]
At each iteration, the algorithm generates $L$ complete episodes according to the current policy, where at the end of each episode, a new environment is sampled from $q$ and reset and the hidden states of the policy and value function are reset.
It keeps the $\epsilon$ fraction of episodes with lowest reward and uses them to update the policy via a vanilla RL algorithm (TRPO~\citep{trpo} in the paper).
We use A2C and PPO, implementing EPOpt on top of them.

\mypara{RL$^2$}.
In order to maximize the expected reward over a distribution of environments, RL$^2$ tries to train an agent that can adapt to the dynamics of the environment at hand.
RL$^2$ models the policy and value functions as a recurrent neural network (RNN) with the current trajectory as input, not just the sequence of states.
The hidden states of the RNN may be viewed as an environment embedding.
Specifically, for the RNN the inputs at time $t$ are $\bs_t$, $\ba_{t-1}$, $r_{t-1}$, and $d_{t-1}$, where $d_{t-1}$ is a Boolean variable indicating whether the episode ended after taking action $\ba_{t-1}$; the output is $\ba_t$ and the hidden states are updated.
Like the other algorithms, at each iteration trajectories are generated using the current policy with the environment state reset at the end of each episode.
However, unlike the other algorithms, a new environment is sampled from $q$ only at the end of every $N$ episodes, which we call a trial.
The generated trajectories are then input into a model-free RL algorithm, maximizing expected reward in a trial; the paper uses TRPO, while we use A2C and PPO\@.
As with EPOpt, our implementation of RL$^2$ is built on top of those of A2C and PPO\@.

\section{Environments}
\label{sec:envs}

\begin{figure}[ht]
  \vskip 0.1in
  \begin{center}
  \includegraphics[width=0.6\columnwidth]{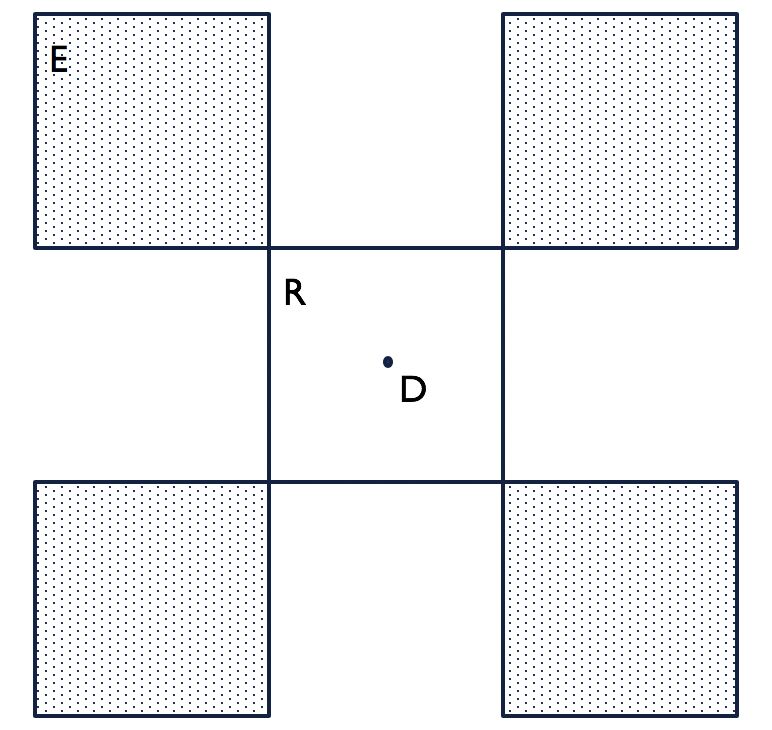}
  \caption{Schematic of the three versions of an environment.}
  \label{fig:DRE}
  \end{center}
  \vskip -0.1in
\end{figure}

Our environments are modified versions of four environments from the classic control problems in OpenAI Gym~\citep{gym} (CartPole, MountainCar, Acrobot, and Pendulum) and two environments from OpenAI Roboschool~\citep{pporoboschool} (HalfCheetah and Hopper) that are based on the corresponding MuJoCo~\citep{mujoco} environments.
We alter the implementations to allow control of several environment parameters that affect the system dynamics, i.e. transition probability distribution functions of the corresponding MDPs.
Similar environments have been used in experiments in the generalization for RL literature, see for instance \cite{epopt}, \cite{pintorobust}, and \cite{yu2017}.
Each of the six environments has three versions, with $d$ parameters allowed to vary.
Figure~\ref{fig:DRE} is a schematic of the parameter ranges in D, R, and E when $d=2$.
\begin{enumerate}
\item Deterministic (D): The parameters of the environment are fixed at the default values in the implementations from Gym and Roboschool. That is, every time the environment is reset, only the state is reset.
\item Random (R): Every time the environment is reset, the parameters are uniformly sampled from a $d$-dimensional box containing the default values. This is done by independently sampling each parameter uniformly from an interval containing the default value.
\item Extreme (E): Every time the environment is reset, its parameters are uniformly sampled from $2^d$ $d$-dimensional boxes anchored at the vertices of the box in R\@. This is done by independently sampling each parameter uniformly from the union of two intervals that straddle the corresponding interval in R\@.
\end{enumerate}

The structure of the three versions was designed to mirror the process of training an RL agent on a real-world task. 
D symbolizes the fixed environment used in the classic RL setting, and R represents the distribution of environments from which it is feasible and sensible to obtain training data.
With more extreme parameters, E corresponds to edge cases, those unusual environments that are not seen during training but must be handled in deployment.

\mypara{CartPole} \citep{cartpole}.
A pole is attached to a cart that moves on a frictionless track.
For at most $200$ time steps, the agent pushes the cart either left or right in order to keep the pole upright.
There is a reward of $1$ for each time step the pole is upright, with the episode ending when the angle of the pole from vertical is too large.
Three environment parameters can be varied: (1) push force magnitude, (2) pole length, and (3) pole mass.

\mypara{MountainCar} \citep{mountaincar}.
The goal is to move a car to the top of a hill within $200$ time steps.
At each time step, the agent pushes a car left or right, with a reward of $-1$. 
Two environment parameters can be varied: (1) push force magnitude and (2) car mass.

\mypara{Acrobot} \citep{suttongeneralization}.
The acrobot is a two-link pendulum attached to a bar with an actuator at the joint between the two links.
For at most $500$ time steps, the agent applies torque (to the left, to the right, or not at all) to the joint in order to swing the end of the second link above the bar to a height equal to the length of the link. 
The reward structure is the same as that of MountainCar.
Three parameters of the links can be varied: (1) length, (2) mass, and (3) moment of inertia, which are the same for each link.

\mypara{Pendulum}.
The goal is to, for $200$ time steps, apply a continuous-valued force to a pendulum in order to keep it at a vertical position.
The reward at each time step is a decreasing function of the pendulum's angle from vertical, the speed of the pendulum, and the magnitude of the applied force.
Two environment parameters can be varied, the pendulum's (1) length and (2) mass.

\mypara{HalfCheetah}.
The half-cheetah is a bipedal robot with eight links and six actuated joints corresponding to the thighs, shins, and feet.
The goal is for the robot to learn to walk on a track without falling over by applying continuous-valued forces to its joints.
The reward at each time step is a combination of the progress made and the costs of the movements, e.g., electricity and penalties for collisions, with a maximum of $1000$ time steps.
Three environment parameters can be varied: (1) power, a factor by which the forces are multiplied before application, (2) torso density, and (3) sliding friction of the joints.

\mypara{Hopper}.
The hopper is a monopod robot with four links arranged in a chain corresponding to a torso, thigh, shin, and foot and three actuated joints.
The goal, reward structure, and varied parameters are the same as those of HalfCheetah.

In all environments, the difficulty may depend on the values of the parameters; for example, in CartPole, a very light and long pole would be more difficult to balance.
By sampling parameters from boxes surrounding the default values, R and E include environments of various difficulties.
The actual ranges of the parameters for each environment, shown in Table~\ref{tab:paramranges}, were chosen by hand so that a policy trained on D struggles in environments in R and fails in E.

\begin{table*}[!htb]
\vskip 0.1in
\begin{center}
\caption{Ranges of parameters for each version of each environment, using set notation.}
\label{tab:paramranges}
\resizebox{0.8\linewidth}{!}{
  \small
  \begin{tabular}{@{}l@{\hspace{6mm}}l@{\hspace{6mm}}c@{\hspace{4mm}}c@{\hspace{6mm}}c@{\hspace{6mm}}}
    \toprule
    Environment & Parameter & D & R & E \\
    \midrule
    CartPole & Force & 10 & [5,15] & [1,5]$\cup$[15,20]   \\
    & Length & 0.5 & [0.25,0.75] & [0.05,0.25]$\cup$[0.75,1.0] \\
    & Mass & 0.1 & [0.05,0.5] & [0.01,0.05]$\cup$[0.5,1.0] \\
    \midrule
    MountainCar & Force & 0.001& [0.0005,0.005] & [0.0001,0.0005]$\cup$[0.005,0.01] \\
    & Mass & 0.0025 & [0.001,0.005] & [0.0005,0.001]$\cup$[0.005,0.01] \\
    \midrule
    Acrobot & Length & 1 & [0.75,1.25] & [0.5,0.75]$\cup$[1.25,1.5] \\
    & Mass & 1 & [0.75,1.25] & [0.5,0.75]$\cup$[1.25,1.5] \\
    & MOI & 1 & [0.75,1.25] & [0.5,0.75]$\cup$[1.25,1.5] \\
    \midrule
    Pendulum & Length & 1 & [0.75,1.25] & [0.5,0.75]$\cup$[1.25,1.5] \\
    & Mass & 1 & [0.75,1.25] & [0.5,0.75]$\cup$[1.25,1.5] \\
    \midrule
    HalfCheetah & Power & 0.90 & [0.70,1.10] & [0.50,0.70]$\cup$[1.10,1.30] \\
    & Density & 1000 & [750,1250] & [500,750]$\cup$[1250,1500] \\
    & Friction & 0.8 & [0.5,1.1] & [0.2,0.5]$\cup$[1.1,1.4] \\
    \midrule
    Hopper & Power & 0.75 & [0.60,0.90] & [0.40,0.60]$\cup$[0.90,1.10] \\
    & Density & 1000 & [750,1250] & [500,750]$\cup$[1250,1500] \\
    & Friction & 0.8 & [0.5,1.1] & [0.2,0.5]$\cup$[1.1,1.4] \\
    \bottomrule
  \end{tabular}
}
\end{center}
\vskip -0.1in
\end{table*}

\subsection{Performance metrics}
\label{sec:metric}

The traditional performance metric used in the RL literature is the average total reward achieved by the policy in an episode.
In the spirit of the definition of an RL agent as goal-seeking~\citep{suttonbarto}, we compute the percentage of episodes in which a certain goal is successfully completed, the success rate.
We define the goals of each environment as follows:
(1) CartPole: balance for at least $195$ time steps,
(2) MountainCar: get to the hilltop within $110$ time steps,
(3) Acrobot: swing the end of the second link to the desired height within $80$ time steps,
(4) Pendulum: keep the angle of the pendulum at most $\pi/3$ radians from vertical for the last $100$ time steps of a trajectory with length $200$, and
(5) HalfCheetah and Hopper: walk for $20$ meters.
The goals for CartPole, MountainCar, and Acrobot are based on the definition of success for those environments given by OpenAI Gym.

The binary success rate is clear and interpretable and has multiple advantages.
First, it is independent of reward shaping, the common practice of modifying an environment's reward functions to help an agent to learn desired behaviors. 
Moreover, separate implementations of the same environment, e.g. HalfCheetah in Roboschool and rllab~\cite{rllab}, may have different reward functions which are hidden in the code and not easily understood. 
Second, it allows fair comparisons across various environment parameters. 
For example, a slightly heavier torso in HalfCheetah and Hopper would change the energy consumption of the robot and thus the reward function in Roboschool but would not change the definition of success in terms of walking a certain distance.

\section{Experimental methodology}
\label{sec:setup}

We benchmark six algorithms (A2C, PPO, EPOpt-A2C, EPOpt-PPO, RL$^2$-A2C, RL$^2$-PPO) and six environments (CartPole, MountainCar, Acrobot, Pendulum, HalfCheetah, Hopper).
With each pair of algorithm and environment, we consider nine training-testing scenarios: training on D, R, and E and testing on D, R, and E\@.
We refer to each scenario using the two-letter abbreviation of the training and testing environment versions, e.g., DR for training on D and testing on R\@.
For A2C, PPO, EPOpt-A2C, and EPOpt-PPO, we train for $15000$ episodes and test on $1000$ episodes.
For RL$^2$-A2C and RL$^2$-PPO, we train for $7500$ trials of $2$ episodes each, equivalent to $15000$ episodes, and test on the last episodes of $1000$ trials; this allows us to evaluate the ability of the policies to adapt to the environment in the current trial.
Note that this is a fair protocol as policies without memory of previous episodes are expected to have the same performance in any episode of a trial.
We do a thorough sweep of hyperparameters, and for the sake of fairness, we randomly generate random seeds and report results over several runs of the entire sweep.
In the following paragraphs we describe the network architectures for the policy and value functions, our hyperparameter search, and the performance metrics we use for evaluation.

\mypara{Evaluation metrics}. For each algorithm, architecture, and environment, we compute three numbers that distill the nine success rates into simple metrics for generalization performance: (1) Default: success percentage on DD, (2) Interpolation: success percentage on RR, and (3) Extrapolation: geometric mean of the success percentages on DR, DE, and RE\@.

Default is the classic RL setting and thus provides a baseline for comparison.
Because R represents the set of environments seen during training in a real-world RL problem, the success rate on RR, or Interpolation, measures the generalization performance to environments similar to those seen during training.
In DR, DE, and RE the training distribution of environments does not overlap with the test distribution. 
Therefore, their success rates are combined to obtain Extrapolation, which measures the generalization performance to environments different from those seen during training.

\mypara{Policy and value function parameterization}. To have equitable evaluations, we consider two network architectures for the policy and value functions for all algorithms and environments.
In the first, following others in the literature including \citet{Henderson2017}, the policy and value functions are multi-layer perceptrons (MLPs) with two hidden layers of $64$ units each and hyperbolic tangent activations; there is no parameter sharing. We refer to this architecture as FF (feed-forward).
In the second,\footnote{Based on personal communication with authors of \citet{rlsquared}.} the policy and value functions are the outputs of two separate fully-connected layers on top of a one-hidden-layer RNN with long short-term memory (LSTM) cells of $256$ units.
The RNN itself is on top of a MLP with two hidden layers of $256$ units each, which we call the feature network. 
Again, hyperbolic tangent activations are used throughout; we refer to this architecture as RC (recurrent).
For A2C, PPO, EPOpt-A2C, and EPOpt-PPO, we evaluate both architectures (where the inputs are the environment states), while for RL$^2$-A2C and RL$^2$-PPO, we evaluate only the second architecture (where the input is a tuple of states, actions, rewards, and Booleans as discussed in Section~\ref{sec:algs}).
In all cases, for discrete action spaces policies sample actions by taking a softmax function over the policy network output layer; for continuous action spaces actions are sampled from a Gaussian distribution with mean the policy network output and diagonal covariance matrix whose entries are learned along with the policy and value function network parameters.

\mypara{Hyperparameters}. During training, for each algorithm and each version of each environment, we performed grid search over a set of hyperparameters used in the optimizers, and selected the value with the highest success probability when tested on the same version of the environment.
That set of hyperparameters includes the learning rate for all algorithms and the length of the trajectory generated at each iteration for A2C, PPO, RL$^2$-A2C, and RL$^2$-PPO\@.
They also include the coefficient of the policy entropy in the objective for A2C, EPOpt-A2C, and RL$^2$-A2C and the coefficient of the KL divergence between the previous policy and current policy for RL$^2$-PPO\@.
The grid values are listed in the supplement.
Other hyperparameters, such as the discount factor, were fixed at the default values in OpenAI Baselines, or in the case of EPOpt-A2C and EPOpt-PPO, $L=100$ and $\epsilon=1.0$ for $100$ iterations and then $\epsilon=0.1$.

\section{Results and discussion}
\label{sec:results}

We highlight some of the key findings and present a summary of the experimental results in Table~\ref{tab:resultssummary}.
The supplement contains analogous tables for each environment, which also informs the following discussion. 

\begin{table*}[tb]
\centering
\caption{Generalization performance (in \% success) of each algorithm, averaged over all environments (mean and standard deviation over five runs).}
\vskip 0.1in
\label{tab:resultssummary}
\newcolumntype{C}{@{}>{${}}c<{{}$}@{} }
\resizebox{0.7\linewidth}{!}{
    \begin{tabular}{l@{\hspace{6mm}}c@{\hspace{6mm}} rCl@{\hspace{6mm}} rCl@{\hspace{6mm}} rCl}
        \toprule
        Algorithm & Architecture & \multicolumn{3}{c}{Default} & \multicolumn{3}{c}{Interpolation} & \multicolumn{3}{c}{Extrapolation} \\
        \midrule
        A2C & FF & 78.14 & \pm & 6.07 & 76.63 & \pm & 1.48 & 63.72 & \pm & 2.08 \\
         & RC & 81.25 & \pm & 3.48 & 72.22 & \pm & 2.95 & 60.76 & \pm & 2.80 \\
        \midrule
        PPO & FF & 78.22 & \pm & 1.53 & 70.57 & \pm & 6.67 & 48.37 & \pm & 3.21 \\
         & RC & 26.51 & \pm & 9.71 & 41.03 & \pm & 6.59 & 21.59 & \pm & 10.08  \\
        \midrule
        EPOpt-A2C & FF & 2.46 & \pm & 2.86 & 7.68 & \pm & 0.61 & 2.35 & \pm & 1.59 \\
         & RC & 9.91 & \pm & 1.12 & 20.89 & \pm & 1.39 & 5.42 & \pm & 0.24 \\
        \midrule
        EPOpt-PPO & FF & 85.40 & \pm & 8.05 & 85.15 & \pm & 6.59 & 59.26 & \pm & 5.81 \\
         & RC & 5.51 & \pm & 5.74 & 15.40 & \pm & 3.86 & 9.99 & \pm & 7.39  \\
        \midrule
        RL$^2$-A2C & RC & 45.79 & \pm & 6.67 & 46.32 & \pm & 4.71 & 33.54 & \pm & 4.64 \\
        \midrule
        RL$^2$-PPO & RC & 22.22 & \pm & 4.46 & 29.93 & \pm & 8.97 & 21.36 & \pm & 4.41 \\
        \bottomrule
    \end{tabular}
}
\vskip -0.1in
\end{table*}

\mypara{A2C and PPO}. We first consider the results under the FF architecture.
The two vanilla deep RL algorithms are usually successful in the classic RL setting of training and testing on a fixed environment, as evidenced by the high values for Default in Table~\ref{tab:resultssummary}.
However, when those agents trained on the fixed environment D are tested, we observed that they usually suffer from a significant drop in performance in R and an even further drop in E\@.
Fortunately, based on the performance in RR, we see that simply training on a distribution of environments, without adding any special mechanism for generalization, results in agents that can perform fairly well in similar environments.
\footnote{We have also found that sometimes the performance in RD is better than that in DD. That is, adding variation to a fixed environment can help to stabilize training enough that the algorithm finds a better policy for the original environment.}
For example, PPO, for which Default equals $78.22$, has Interpolation equal to $70.57$.
Nevertheless, as expected in general they are less successful at extrapolation; PPO has Extrapolation equal to $48.37$.
With the RC architecture, A2C has similar behavior as with the FF architecture while PPO had difficulty training on the fixed environment D and did not generalize well.
For example, on all the environments except CartPole and Pendulum the FF architecture was necessary for PPO to be successful even in the classic RL setting.

The pattern of declining performance from Default to Interpolation to Extrapolation also appears when looking at each environment individually.
The magnitude of decrease depends on the combination of algorithm, architecture, and environment.
For instance, on CartPole, A2C interpolates and extrapolates successfully, where Interpolation equals $100.00$ and Extrapolation equals $93.63$ with the FF architecture and $83.00$ with the RC architecture; this behavior is also shown for PPO with the FF architecture\@. 
On the other hand, on Hopper, PPO with the FF architecture has $85.54\%$ success rate in the classic RL setting but struggles to interpolate (Interpolation equals $39.68$) and fails to extrapolate (Extrapolation equals $10.36$).
This indicates that our choices of environments and their parameter ranges lead to a variety of difficulty in generalization.

\mypara{EPOpt}. Again, we start with the results under the FF architecture. 
Overall EPOpt-PPO improved both interpolation and extrapolation performance over PPO\@, as shown in Table~\ref{tab:resultssummary}.
Looking at specific environments, on Hopper EPOpt-PPO has nearly twice the interpolation performance and significantly improved  extrapolation performance compared to PPO\@. 
Such an improvement also appears for Pendulum and HalfCheetah.
However, the generalization performance of EPOpt-PPO was worse than that of PPO on the other three environments.
EPOpt did not demonstrate the same performance gains when combined with A2C\@; in fact, it generally failed to train even in the fixed environment D.
With the RC architecture, EPOpt-PPO, like PPO, also had difficulties training on environment D.
EPOpt-A2C was able to find limited success on CartPole but failed to learn a working policy in environment D for the other environments.
The effectiveness of EPOpt when combined with PPO but not A2C and then only on Pendulum, Hopper, and HalfCheetah indicates that a continuous action space is important for its success.

\mypara{RL$^2$}. RL$^2$-A2C and RL$^2$-PPO proved to be difficult to train and data inefficient.
On most environments, the Default numbers are low, indicating that a working policy was not found in the classic RL setting of training and testing on a fixed environment.
As a result, they also have low Interpolation and Extrapolation numbers.
This suggests that additional structure must be injected into the policy in order to learn useful environmental characteristics from the trajectories. 
The difficulty in training may also be partially due to the recurrent architecture, as PPO with the RC architecture does not find a successful policy in the classic RL setting as shown in Table~\ref{tab:resultssummary}.
Thus, using a Temporal Convolution Network may offer improvements, as they have been shown to outperform RNNs in many sequence modeling tasks~\citep{bai2018empirical}.
The same qualitative observations also hold for other values of the number of episodes per trial; see the supplement for specifics.

In a few environments, such as RL$^2$-PPO on CartPole and RL$^2$-A2C on HalfCheetah, a working policy was found in the classic RL setting, but the algorithm struggled to interpolate or extrapolate.
The one success story is RL$^2$-A2C on Pendulum, where we have nearly $100\%$ success rate in DD, interpolate extremely well (Interpolation is $99.82$), and extrapolate fairly well (Extrapolation is $81.79$).
We observed that the partial success of these algorithms on the environments appears to be dependent on two implementation choices: the feature network in the RC architecture and the nonzero coefficient of the KL divergence between the previous policy and current policy in RL$^2$-PPO, which is intended to help stabilize training.

\vspace{-1mm}
\section{Conclusion}
\label{sec:conclusion}

We presented an empirical study of generalization in deep RL.
We evaluated two state-of-the-art deep RL algorithms, A2C and PPO, and two algorithms that explicitly tackle the problem of generalization in different ways: EPOpt, which aims to be robust to environment variations, and RL$^2$, which aims to automatically adapt to them.
In order to do this, we introduced a new testbed and experimental protocol to measure the ability of RL algorithms to generalize to environments both similar to and different from those seen during training.
A common testbed and protocol, which were missing from previous work, enable us to compare the relative merits of algorithms for building generalizable RL agents.
\ifisaccepted
Our code is available online~\footnote{\href{http://www.github.com/sunblaze-ucb/rl-generalization}{http://www.github.com/sunblaze-ucb/rl-generalization}} and we hope that it will support future research on generalization in deep RL\@.
\else
Our code will be made available online and we hope that it will support future research on generalization in deep RL\@.
\fi

Overall, the vanilla deep RL algorithms have better generalization performance than their more complex counterparts, being able to interpolate quite well with some extrapolation success.
In other words, vanilla deep RL algorithms trained with environmental stochasticity may be more effective for generalization than specialized algorithms; the same conclusion was also suggested by the results of the OpenAI Retro contest~\citep{openairetro} and the CoinRun benchmark~\citep{coinrun} in environments with visual input.
When combined with PPO under the FF architecture, EPOpt is able to outperform PPO, in particular for the environments with continuous action spaces; however, it does not generalize in the other cases and often fails to train even on the Default environments.
RL$^2$ is difficult to train, and in its success cases provides no clear generalization advantage over the vanilla deep RL algorithms or EPOpt.

The sensitivity of the effectiveness of EPOpt and RL$^2$ to the base algorithm, architecture, and environment presents an avenue for future work, as EPOpt and RL$^2$ were presented as general-purpose techniques.
We have considered model-free RL algorithms in our evaluation; another direction for future work is to further investigate model-based RL algorithms, such as the recent work of \citet{saemundsson2018} and \citet{clavera2018}.
Because model-based RL explicitly learns the system dynamics and is generally more data efficient, it could be better leveraged by adaptive techniques for generalization.

\bibliography{main}
\bibliographystyle{icml2019}

\ifisaccepted
\section*{Acknowledgments}
This material is in part based upon work supported by the National Science Foundation under 
Grant No. TWC-1409915,
DARPA under FA8750-17-2-0091,
and Berkeley Deep Drive.
Any opinions, findings, and conclusions or recommendations expressed in this material are those of the authors and do not necessarily reflect the views of the National Science Foundation.
\else
\fi

\appendix

\section{Training Hyperparameters}
\label{sec:hyperparameters}

The grid values we search over for each hyperparameter and each algorithm are listed below. In sum, the search space contains $183$ unique hyperparameter configurations for all algorithms on a single training environment ($3,294$ training configurations), and each trained agent is evaluated on $3$ test settings ($9,882$ total train/test configurations).
We report results for $5$ runs of the full grid search, a total of $49,410$ experiments.
\begin{itemize}
    \item Learning rate:
    \begin{itemize}
        \item A2C, EPOpt-A2C with RC architecture, and RL$^2$-A2C: [$7\mathrm{e}{-3}$, $7\mathrm{e}{-4}$, $7\mathrm{e}{-5}$]
        \item EPOpt-A2C with FF architecture: [$7\mathrm{e}{-2}$, $7\mathrm{e}{-3}$, $7\mathrm{e}{-4}$]
        \item PPO, EPOpt-PPO with RC architecture: [$3\mathrm{e}{-3}$, $3\mathrm{e}{-4}$, $3\mathrm{e}{-5}$]
        \item EPOpt-PPO with FF architecture: [$3\mathrm{e}{-2}$, $3\mathrm{e}{-3}$, $3\mathrm{e}{-4}$]
        \item RL$^2$-PPO: [$3\mathrm{e}{-4}$, $3\mathrm{e}{-5}$, $3\mathrm{e}{-6}$]
    \end{itemize}
    \item Length of the trajectory generated at each iteration:
    \begin{itemize}
        \item A2C and RL$^2$-A2C: $[5, 10, 15]$
        \item PPO and RL$^2$-PPO: $[128, 256, 512]$
    \end{itemize}
    \item Policy entropy coefficient: $[1\mathrm{e}{-2}, 1\mathrm{e}{-3}, 1\mathrm{e}{-4}, 1\mathrm{e}{-5}]$
    \item KL divergence coefficient: $[0.3, 0.2, 0.0]$
\end{itemize}

\section{Detailed Experimental Results}
\label{sec:detailedresults}

In order to elucidate the generalization behavior of each algorithm, here we present the quantities in Table $1$ of the paper for each environment.

\begin{table*}[!htb]
\vskip 0.1in
\centering
\setlength{\tabcolsep}{1mm}
\caption{Mean and standard deviation over five runs of generalization performance (in \% success) on Acrobot.}\vspace{2mm}
\label{tab:resultsAcrobot}
\newcolumntype{C}{@{}>{${}}c<{{}$}@{} }
\resizebox{0.75\linewidth}{!}{
    \begin{tabular}{l@{\hspace{6mm}}c@{\hspace{6mm}} rCl@{\hspace{6mm}} rCl@{\hspace{6mm}} rCl}
        \toprule
        Algorithm & Architecture & \multicolumn{3}{c}{Default} & \multicolumn{3}{c}{Interpolation} & \multicolumn{3}{c}{Extrapolation} \\
        \midrule
        A2C & FF & 88.52 & \pm & 1.32 & 72.88 & \pm & 0.74 & 66.56 & \pm & 0.52 \\
         & RC & 88.24 & \pm & 1.53 & 73.46 & \pm & 1.11 & 67.94 & \pm & 1.06 \\
        \midrule
        PPO & FF & 87.20 & \pm & 1.11 & 72.78 & \pm & 0.44 & 64.93 & \pm & 1.05 \\
         & RC & 0.0 & \pm & 0.0 & 0.0 & \pm & 0.0 & 0.0 & \pm & 0.0  \\
        \midrule
        EPOpt-A2C & FF & 0.0 & \pm & 0.0 & 0.0 & \pm & 0.0 & 0.0 & \pm & 0.0 \\
         & RC & 0.0 & \pm & 0.0 & 0.0 & \pm & 0.0 & 0.0 & \pm & 0.0 \\
        \midrule
        EPOpt-PPO & FF & 79.60 & \pm & 5.86 & 69.20 & \pm & 1.64 & 65.05 & \pm & 2.16 \\
         & RC & 3.10 & \pm & 3.14 & 6.40 & \pm & 3.65 & 15.57 & \pm & 5.59  \\
        \midrule
        RL$^2$-A2C & RC & 65.70 & \pm & 8.68 & 57.70 & \pm & 2.40 & 57.01 & \pm & 2.70 \\
        \midrule
        RL$^2$-PPO & RC & 0.0 & \pm & 0.0 & 0.0 & \pm & 0.0 & 0.0 & \pm & 0.0 \\
        \bottomrule
    \end{tabular}
}
\vskip -0.1in
\end{table*}

\begin{table*}[!htb]
\vskip 0.1in
\centering
\setlength{\tabcolsep}{1mm}
\caption{Mean and standard deviation over five runs of generalization performance (in \% success) on CartPole.}\vspace{2mm}
\label{tab:resultsCartPole}
\newcolumntype{C}{@{}>{${}}c<{{}$}@{} }
\resizebox{0.75\linewidth}{!}{
    \begin{tabular}{l@{\hspace{6mm}}c@{\hspace{6mm}} rCl@{\hspace{6mm}} rCl@{\hspace{6mm}} rCl}
        \toprule
        Algorithm & Architecture & \multicolumn{3}{c}{Default} & \multicolumn{3}{c}{Interpolation} & \multicolumn{3}{c}{Extrapolation} \\
        \midrule
        A2C & FF & 100.00 & \pm & 0.0 & 100.00 & \pm & 0.0 & 93.63 & \pm & 9.30 \\
         & RC & 100.00 & \pm & 0.0 & 100.00 & \pm & 0.0 & 83.00 & \pm & 11.65 \\
        \midrule
        PPO & FF & 100.00 & \pm & 0.0 & 100.00 & \pm & 0.0 & 86.20 & \pm & 12.60 \\
         & RC & 65.58 & \pm & 27.81 & 70.80 & \pm & 21.02 & 45.00 & \pm & 18.06  \\
        \midrule
        EPOpt-A2C & FF & 14.74 & \pm & 17.14 & 43.06 & \pm & 3.48 & 10.48 & \pm & 9.41 \\
         & RC & 57.00 & \pm & 4.50 & 55.88 & \pm & 3.97 & 32.53 & \pm & 1.47 \\
        \midrule
        EPOpt-PPO & FF & 99.98 & \pm & 0.04 & 99.46 & \pm & 0.79 & 73.58 & \pm & 12.19 \\
         & RC & 29.94 & \pm & 31.58 & 20.22 & \pm & 17.83 & 14.55 & \pm & 20.09  \\
        \midrule
        RL$^2$-A2C & RC & 20.78 & \pm & 39.62 & 0.06 & \pm & 0.12 & 0.12 & \pm & 0.23 \\
        \midrule
        RL$^2$-PPO & RC & 87.20 & \pm & 12.95 & 54.22 & \pm & 34.85 & 51.00 & \pm & 14.60 \\
        \bottomrule
    \end{tabular}
}
\vskip -0.1in
\end{table*}

\begin{table*}[!htb]
\vskip 0.1in
\centering
\setlength{\tabcolsep}{1mm}
\caption{Mean and standard deviation over five runs of generalization performance (in \% success) on MountainCar.}\vspace{2mm}
\label{tab:resultsMountainCar}
\newcolumntype{C}{@{}>{${}}c<{{}$}@{} }
\resizebox{0.75\linewidth}{!}{
        \begin{tabular}{l@{\hspace{6mm}}c@{\hspace{6mm}} rCl@{\hspace{6mm}} rCl@{\hspace{6mm}} rCl}
        \toprule
        Algorithm & Architecture & \multicolumn{3}{c}{Default} & \multicolumn{3}{c}{Interpolation} & \multicolumn{3}{c}{Extrapolation} \\
        \midrule
        A2C & FF & 79.78 & \pm & 11.38 & 84.10 & \pm & 1.25 & 89.72 & \pm & 0.65 \\
         & RC & 95.88 & \pm & 4.10 & 74.84 & \pm & 6.82 & 89.77 & \pm & 0.76 \\
        \midrule
        PPO & FF & 99.96 & \pm & 0.08 & 84.12 & \pm & 0.84 & 90.21 & \pm & 0.37 \\
         & RC & 0.0 & \pm & 0.0 & 63.36 & \pm & 0.74 & 15.86 & \pm & 31.71  \\
        \midrule
        EPOpt-A2C & FF & 0.0 & \pm & 0.0 & 3.04 & \pm & 0.19 & 3.63 & \pm & 0.49 \\
         & RC & 0.0 & \pm & 0.0 & 62.46 & \pm & 0.80 & 0.0 & \pm & 0.0 \\
        \midrule
        EPOpt-PPO & FF & 74.42 & \pm & 37.93 & 84.86 & \pm & 1.09 & 87.42 & \pm & 5.11 \\
         & RC & 0.0 & \pm & 0.0 & 65.74 & \pm & 4.88 & 29.82 & \pm & 27.30  \\
        \midrule
        RL$^2$-A2C & RC & 0.32  & \pm & 0.64 & 57.86 & \pm & 2.97 & 21.56 & \pm & 30.35 \\
        \midrule
        RL$^2$-PPO & RC & 0.0 & \pm & 0.0 & 60.10 & \pm & 0.91 & 31.27 & \pm & 26.24 \\
        \bottomrule
    \end{tabular}
}
\vskip -0.1in
\end{table*}

\begin{table*}[!htb]
\vskip 0.1in
\centering
\setlength{\tabcolsep}{1mm}
\caption{Mean and standard deviation over five runs of generalization performance (in \% success) on Pendulum.}\vspace{2mm}
\label{tab:resultsPendulum}
\newcolumntype{C}{@{}>{${}}c<{{}$}@{} }
\resizebox{0.75\linewidth}{!}{
    \begin{tabular}{l@{\hspace{6mm}}c@{\hspace{6mm}} rCl@{\hspace{6mm}} rCl@{\hspace{6mm}} rCl}
        \toprule
        Algorithm & Architecture & \multicolumn{3}{c}{Default} & \multicolumn{3}{c}{Interpolation} & \multicolumn{3}{c}{Extrapolation} \\
        \midrule
        A2C & FF & 100.00 & \pm & 0.0 & 99.86 & \pm & 0.14 & 90.27 & \pm & 3.07 \\
         & RC & 100.00 & \pm & 0.0 & 99.96 & \pm & 0.05 & 79.58 & \pm & 6.41 \\
        \midrule
        PPO & FF & 0.0 & \pm & 0.0 & 31.80 & \pm & 40.11 & 0.0 & \pm & 0.0 \\
         & RC & 73.28 & \pm & 36.80 & 90.94 & \pm & 7.79 & 61.11 & \pm & 31.08  \\
        \midrule
        EPOpt-A2C & FF & 0.0 & \pm & 0.0 & 0.0 & \pm & 0.0 & 0.0 & \pm & 0.0 \\
         & RC & 2.48 & \pm & 4.96 & 7.00 & \pm & 10.81 & 0.0 & \pm & 0.0 \\
        \midrule
        EPOpt-PPO & FF & 100.00 & \pm & 0.0 & 77.34 & \pm & 38.85 & 54.72 & \pm & 27.57 \\
         & RC & 0.0 & \pm & 0.0 & 0.04 & \pm & 0.08 & 0.0 & \pm & 0.0  \\
        \midrule
        RL$^2$-A2C & RC & 100.00 & \pm & 0.0 & 99.82 & \pm & 0.31 & 81.79 & \pm & 3.88 \\
        \midrule
        RL$^2$-PPO & RC & 46.14 & \pm & 17.67 & 65.22 & \pm & 21.78 & 45.76 & \pm & 8.38 \\
        \bottomrule
    \end{tabular}
}
\vskip -0.1in
\end{table*}

\begin{table*}[!htb]
\vskip 0.1in
\centering
\setlength{\tabcolsep}{1mm}
\caption{Mean and standard deviation over five runs of generalization performance (in \% success) on HalfCheetah.}\vspace{2mm}
\label{tab:resultsHalfCheetah}
\newcolumntype{C}{@{}>{${}}c<{{}$}@{} }
\resizebox{0.75\linewidth}{!}{
    \begin{tabular}{l@{\hspace{6mm}}c@{\hspace{6mm}} rCl@{\hspace{6mm}} rCl@{\hspace{6mm}} rCl}
        \toprule
        Algorithm & Architecture & \multicolumn{3}{c}{Default} & \multicolumn{3}{c}{Interpolation} & \multicolumn{3}{c}{Extrapolation} \\
        \midrule
        A2C & FF & 85.06 & \pm & 19.68 & 91.96 & \pm & 8.60 & 40.54 & \pm & 8.34 \\
         & RC & 88.06 & \pm & 12.26 & 74.70 & \pm & 13.49 & 42.96 & \pm & 7.79 \\
        \midrule
        PPO & FF & 96.62 & \pm & 3.84 & 95.02 & \pm & 2.96 & 38.51 & \pm & 15.13 \\
         & RC & 20.22 & \pm & 17.01 & 21.08 & \pm & 26.04 & 7.55 & \pm & 5.04  \\
        \midrule
        EPOpt-A2C & FF & 0.0 & \pm & 0.0 & 0.0 & \pm & 0.0 & 0.0 & \pm & 0.0 \\
         & RC & 0.0 & \pm & 0.0 & 0.0 & \pm & 0.0 & 0.0 & \pm & 0.0 \\
        \midrule
        EPOpt-PPO & FF & 99.76 & \pm & 0.08 & 99.28 & \pm & 0.87 & 53.41 & \pm & 9.41 \\
         & RC & 0.0 & \pm & 0.0 & 0.0 & \pm & 0.0 & 0.0 & \pm & 0.0  \\
        \midrule
        RL$^2$-A2C & RC & 87.96 & \pm & 4.21 & 62.48 & \pm & 29.18 & 40.78 & \pm & 5.99 \\
        \midrule
        RL$^2$-PPO & RC & 0.0 & \pm & 0.0 & 0.0 & \pm & 0.0 & 0.16 & \pm & 0.32 \\
        \bottomrule
    \end{tabular}
}
\vskip -0.1in
\end{table*}

\begin{table*}[!htb]
\vskip 0.1in
\centering
\setlength{\tabcolsep}{1mm}
\caption{Mean and standard deviation over five runs of generalization performance (in \% success) on Hopper.}\vspace{2mm}
\label{tab:resultsHopper}
\newcolumntype{C}{@{}>{${}}c<{{}$}@{} }
\resizebox{0.75\linewidth}{!}{
    \begin{tabular}{l@{\hspace{6mm}}c@{\hspace{6mm}} rCl@{\hspace{6mm}} rCl@{\hspace{6mm}} rCl}
        \toprule
        Algorithm & Architecture & \multicolumn{3}{c}{Default} & \multicolumn{3}{c}{Interpolation} & \multicolumn{3}{c}{Extrapolation} \\
        \midrule
        A2C & FF & 15.46 & \pm & 7.58 & 11.00 & \pm & 7.01 & 1.63 & \pm & 2.77 \\
         & RC & 15.34 & \pm & 8.82 & 10.38 & \pm & 15.14 & 1.31 & \pm & 1.23 \\
        \midrule
        PPO & FF & 85.54 & \pm & 6.96 & 39.68 & \pm & 16.69 & 10.36 & \pm & 6.79 \\
         & RC & 0.0 & \pm & 0.0 & 0.0 & \pm & 0.0 & 0.0 & \pm & 0.0  \\
        \midrule
        EPOpt-A2C & FF & 0.0 & \pm & 0.0 & 0.0 & \pm & 0.0 & 0.0 & \pm & 0.0 \\
         & RC & 0.0 & \pm & 0.0 & 0.0 & \pm & 0.0 & 0.0 & \pm & 0.0 \\
        \midrule
        EPOpt-PPO & FF & 58.62 & \pm & 47.51 & 80.78 & \pm & 29.18 & 21.39 & \pm & 16.62 \\
         & RC & 0.0 & \pm & 0.0 & 0.0 & \pm & 0.0 & 0.0 & \pm & 0.0  \\
        \midrule
        RL$^2$-A2C & RC & 0.0 & \pm & 0.0 & 0.0 & \pm & 0.0 & 0.0 & \pm & 0.0 \\
        \midrule
        RL$^2$-PPO & RC & 0.0 & \pm & 0.0 & 0.02 & \pm & 0.04 & 0.0 & \pm & 0.0 \\
        \bottomrule
    \end{tabular}
}
\vskip -0.1in
\end{table*}

\section{Behavior of MountainCar}
\label{sec:mountaincar}

\begin{figure}[ht]
  \vskip 0.1in
  \begin{center}
  \includegraphics[width=\columnwidth]{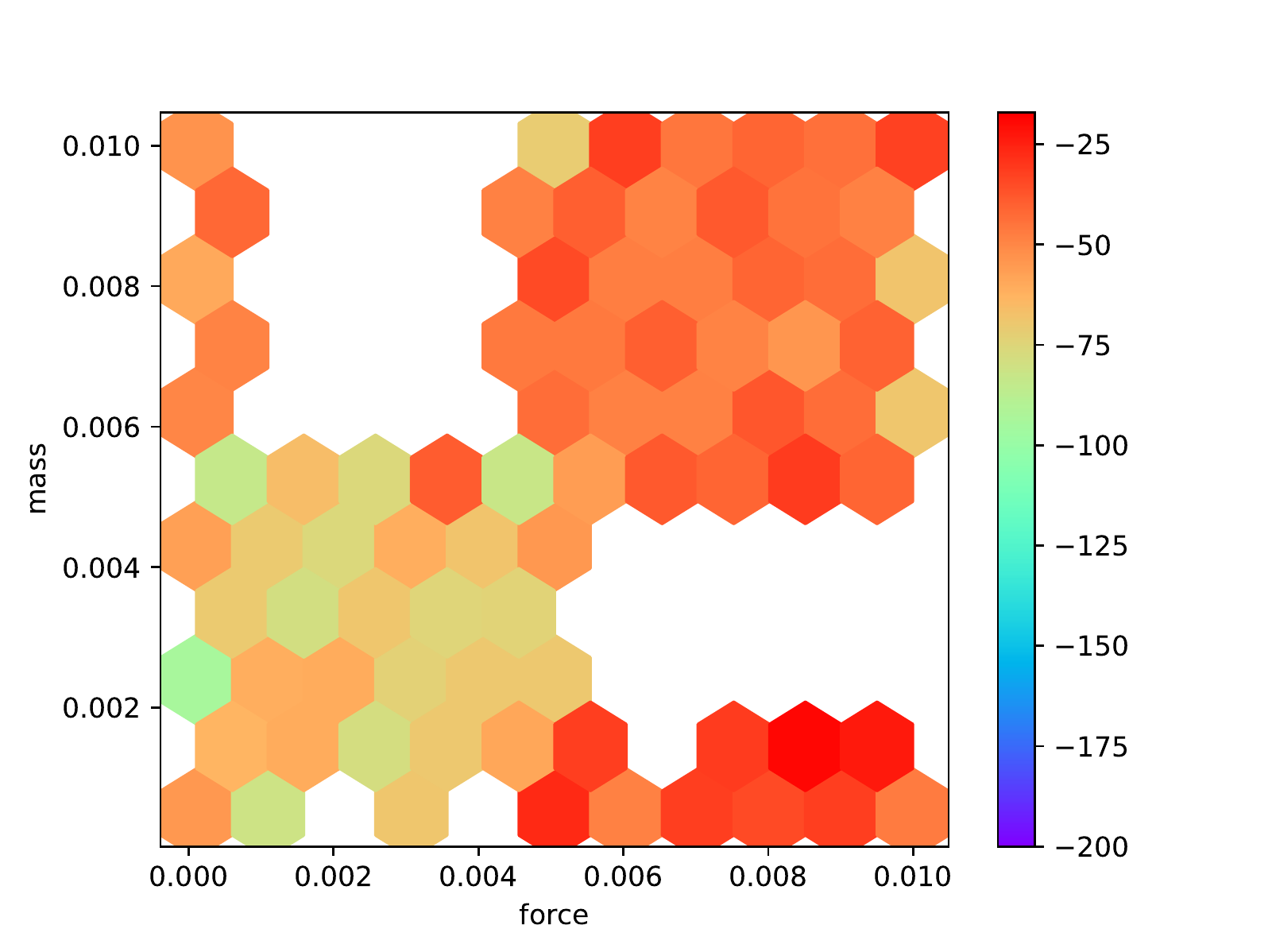}
  \caption{MountainCar: heatmap of the rewards achieved by A2C with the FF architecture on DR and DE\@. The axes are the two environment parameters varied in R and E\@.}
  \label{fig:a2cMCheatmap}
  \end{center}
  \vskip -0.1in
\end{figure}

On MountainCar, several of the algorithms, including A2C with both architectures and PPO with the FF architecture, have greater success on Extrapolation than Interpolation, which is itself sometimes greater than Default (see Table~\ref{tab:resultsMountainCar}).
At first glance, this is unexpected because Extrapolation combines the success rates of DR, DE, and RE, with E containing more extreme parameter settings, while Interpolation is the success rate of RR\@.
To explain this phenomenon, we hypothesize that compared to R, E is dominated by easy parameter settings, e.g., those where the car is light but the force of the push is strong, allowing the agent to reach the top of the hill in only a few steps.
In order to test this hypothesis, we create heatmaps of the rewards achieved by A2C with both architectures and PPO with the FF architecture trained on D and tested on R and E\@. 
We show only the heatmap for A2C with the FF architecture, in Figure~\ref{fig:a2cMCheatmap}; the other two are qualitatively similar.
Referring to the description of the environments in the main paper, we see that the reward achieved by the policy is higher in the regions corresponding to E\@.
Indeed, it appears that the largest regions of E are those with a large force, which enables the trained policy to push the car up the hill in much less than $110$ time steps, achieving the goal defined in Section $6$ of the paper. (Note that the reward is the negative of the number of time steps taken to push the car up the hill.)

\begin{figure}[ht]
  \vskip 0.1in
  \begin{center}
  \includegraphics[width=\columnwidth]{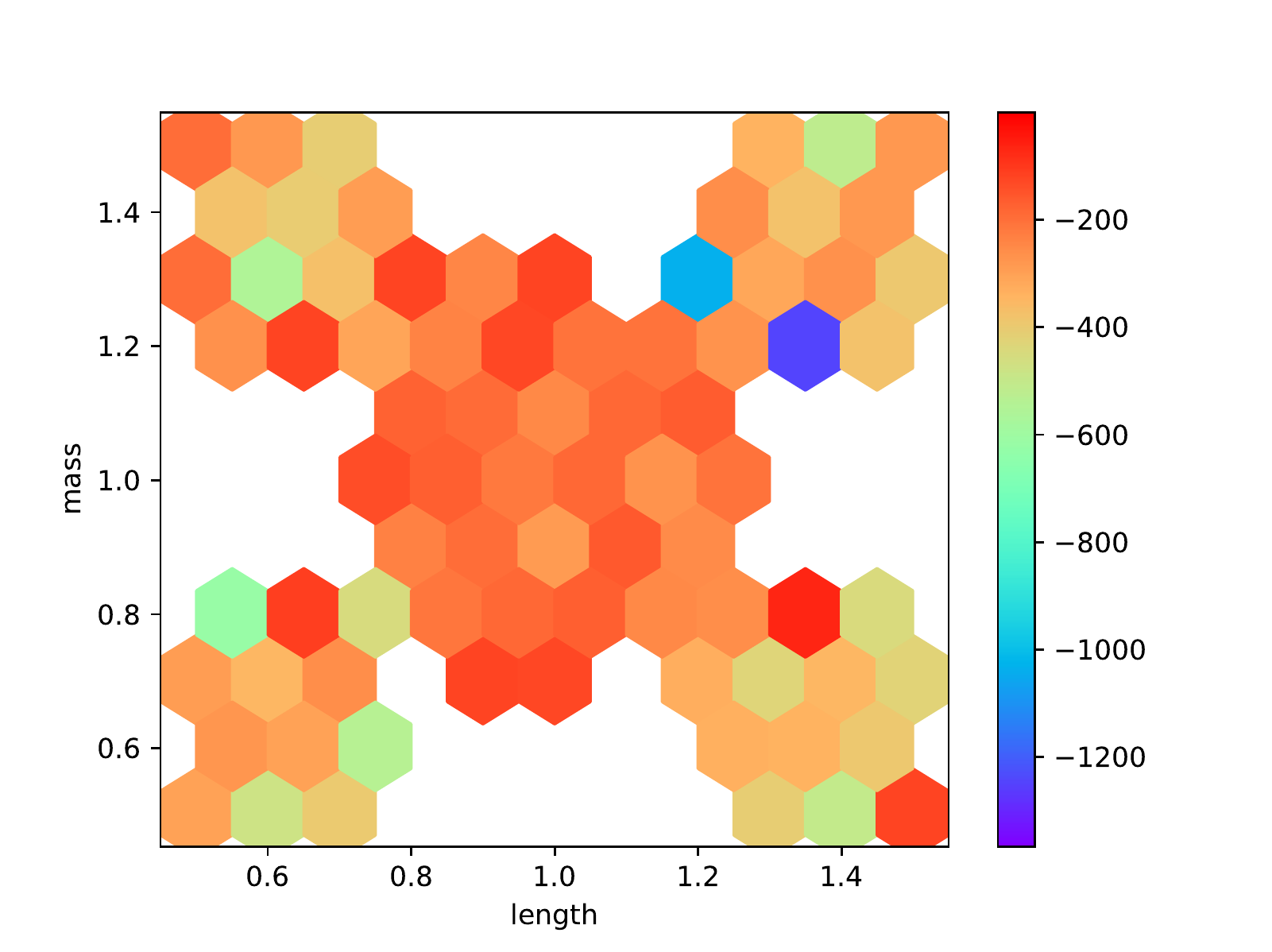}
  \caption{Pendulum: heatmap of the rewards achieved by A2C with the FF architecture on DR and DE\@. The axes are the two environment parameters varied in R and E\@.}
  \label{fig:a2cPendulumheatmap}
  \end{center}
  \vskip -0.1in
\end{figure}

On the other hand, Figure~\ref{fig:a2cPendulumheatmap} shows a similar heatmap for A2C with the FF architecture on Pendulum, in which Interpolation is greater than Extrapolation.
In this case, the policy trained on D struggles more on environments from E than on those from R\@.
This special case demonstrates the importance of considering a wide variety of environments when assessing the generalization performance of an algorithm; each environment may have idiosyncrasies that cause performance to be correlated with parameters.

\section{Varying $N$ in RL$^2$}
\label{sec:varyingN}

We test the sensitivity of the results for RL$^2$-A2C and RL$^2$-PPO to the hyperparameter of the number of episodes per trial, $N$.
We consider $N=1$ and $N=5$, to determine whether it was too difficult to train a RNN over trajectories of $N=2$ episodes or more episodes were needed for adaptation.
Table~\ref{tab:varyingN1} contains the results for $N=1$ on each environment (and averaged); Table~\ref{tab:varyingN5} is a similar table for $N=5$.

It appears that increasing $N$ usually degrades generalization performance, indicating that the increasing trajectory length does make training more difficult.
Nevertheless, there are two special cases where generalization performance improves as $N$ increases, on Acrobot with RL$^2$-A2C and Pendulum with RL$^2$-PPO.

Note that when $N=1$, RL$^2$-A2C is the same as A2C with the RC architecture but with the actions, rewards, and done flags input in addition to the states; the same is true of RL$^2$-PPO and PPO with the RC architecture.
However, on average its generalization performance is not as good; for example Interpolation is $66.83$ for RL$^2$-A2C but $72.22$ for A2C with the RC architecture.
This suggests that RL$^2$ is unable to effectively utilize the information contained in the trajectories to learn about the environment dynamics and a different policy architecture from a simple RNN is needed.

\begin{table*}[!htb]
\vskip 0.1in
\centering
\setlength{\tabcolsep}{1mm}
\caption{Mean and standard deviation over five runs of generalization performance (in \% success) for RL$^2$-A2C and RL$^2$-PPO when $N=1$.}\vspace{2mm}
\label{tab:varyingN1}
\newcolumntype{C}{@{}>{${}}c<{{}$}@{} }
\resizebox{0.75\linewidth}{!}{
    \begin{tabular}{l@{\hspace{6mm}}c@{\hspace{6mm}} rCl@{\hspace{6mm}} rCl@{\hspace{6mm}} rCl}
        \toprule
        Environment & Algorithm & \multicolumn{3}{c}{Default} & \multicolumn{3}{c}{Interpolation} & \multicolumn{3}{c}{Extrapolation} \\
        \midrule
        Average & RL$^2$-A2C & 66.83 & \pm & 7.32 & 65.19 & \pm & 2.86 & 55.76 & \pm & 4.41 \\
        & RL$^2$-PPO & 27.01 & \pm & 8.20 & 34.02 & \pm & 5.92 & 17.59 & \pm & 6.27 \\
        \midrule
        Acrobot & RL$^2$-A2C & 30.54 & \pm & 17.06 & 20.10 & \pm & 16.10 & 33.95 & \pm & 9.43 \\
        & RL$^2$-PPO & 0.0 & \pm & 0.0 & 0.0 & \pm & 0.0 & 0.0 & \pm & 0.0 \\
        \midrule
        CartPole & RL$^2$-A2C & 99.90 & \pm & 0.20 & 100.0 & \pm & 0.0 & 87.50 & \pm & 3.81 \\
        & RL$^2$-PPO & 81.76 & \pm & 22.33 & 86.62 & \pm & 9.47 & 59.10 & \pm & 21.36 \\
        \midrule
        MountainCar & RL$^2$-A2C & 88.94 & \pm & 5.30 & 81.84 & \pm & 0.78 & 86.82 & \pm & 2.62 \\
        & RL$^2$-PPO & 0.0 & \pm & 0.0 & 55.02 & \pm & 0.89 & 11.11 & \pm & 22.23 \\
        \midrule
        Pendulum & RL$^2$-A2C & 100.0 & \pm & 0.0 & 99.98 & \pm & 0.04 & 87.81 & \pm & 2.74 \\
        & RL$^2$-PPO & 42.58 & \pm & 34.73 & 48.80 & \pm & 21.42 & 34.10 & \pm & 24.76 \\
        \midrule
        HalfCheetah & RL$^2$-A2C & 80.54 & \pm & 30.70 & 82.50 & \pm & 14.23 & 35.24 & \pm & 21.03 \\
        & RL$^2$-PPO & 37.70 & \pm & 30.62 & 13.68 & \pm & 15.81 & 1.20 & \pm & 1.49 \\
        \midrule
        Hopper & RL$^2$-A2C & 1.04 & \pm & 1.23 & 6.74 & \pm & 5.71 & 3.26 & \pm & 3.75 \\
        & RL$^2$-PPO & 0.0 & \pm & 0.0 & 0.0 & \pm & 0.0 & 0.0 & \pm & 0.0 \\
        \bottomrule
    \end{tabular}
}
\vskip -0.1in
\end{table*}

\begin{table*}[!htb]
\vskip 0.1in
\centering
\setlength{\tabcolsep}{1mm}
\caption{Mean and standard deviation over five runs of generalization performance (in \% success) for RL$^2$-A2C and RL$^2$-PPO when $N=5$.}\vspace{2mm}
\label{tab:varyingN5}
\newcolumntype{C}{@{}>{${}}c<{{}$}@{} }
\resizebox{0.75\linewidth}{!}{
    \begin{tabular}{l@{\hspace{6mm}}c@{\hspace{6mm}} rCl@{\hspace{6mm}} rCl@{\hspace{6mm}} rCl}
        \toprule
        Environment & Algorithm & \multicolumn{3}{c}{Default} & \multicolumn{3}{c}{Interpolation} & \multicolumn{3}{c}{Extrapolation} \\
        \midrule
        Average & RL$^2$-A2C & 41.08 & \pm & 4.66 & 41.11 & \pm & 2.82 & 27.97 & \pm & 2.37 \\
        & RL$^2$-PPO & 12.77 & \pm & 2.94 & 21.16 & \pm & 1.27 & 15.59 & \pm & 5.39 \\
        \midrule
        Acrobot & RL$^2$-A2C & 79.02 & \pm & 2.67 & 69.86 & \pm & 0.90 & 64.59 & \pm & 0.72 \\
        & RL$^2$-PPO & 0.0 & \pm & 0.0 & 0.0 & \pm & 0.0 & 0.0 & \pm & 0.0 \\
        \midrule
        CartPole & RL$^2$-A2C & 0.0 & \pm & 0.0 & 0.0 & \pm & 0.0 & 0.0 & \pm & 0.0 \\
        & RL$^2$-PPO & 1.04 & \pm & 1.63 & 1.04 & \pm & 1.51 & 0.16 & \pm & 0.31 \\
        \midrule
        MountainCar & RL$^2$-A2C & 0.0 & \pm & 0.0 & 56.60 & \pm & 2.59 & 4.39 & \pm & 2.26 \\
        & RL$^2$-PPO & 0.0 & \pm & 0.0 & 62.32 & \pm & 1.24 & 31.33 & \pm & 27.67 \\
        \midrule
        Pendulum & RL$^2$-A2C & 99.98 & \pm & 0.04 & 99.68 & \pm & 0.35 & 82.67 & \pm & 3.64 \\
        & RL$^2$-PPO & 75.58 & \pm & 19.00 & 63.60 & \pm & 9.27 & 62.07 & \pm & 5.68 \\
        \midrule
        HalfCheetah & RL$^2$-A2C & 67.50 & \pm & 29.37 & 20.50 & \pm & 17.01 & 16.16 & \pm & 11.05 \\
        & RL$^2$-PPO & 0.0 & \pm & 0.0 & 0.0 & \pm & 0.0 & 0.0 & \pm & 0.0 \\
        \midrule
        Hopper & RL$^2$-A2C & 0.0 & \pm & 0.0 & 0.0 & \pm & 0.0 & 0.0 & \pm & 0.0 \\
        & RL$^2$-PPO & 0.0 & \pm & 0.0 & 0.0 & \pm & 0.0 & 0.0 & \pm & 0.0 \\
        \bottomrule
    \end{tabular}
}
\vskip -0.1in
\end{table*}

\section{Training Curves}
\label{sec:trainingcurves}

\begin{figure*}[ht]
    \subfigure[PPO with FF architecture]{\includegraphics[width=\columnwidth]{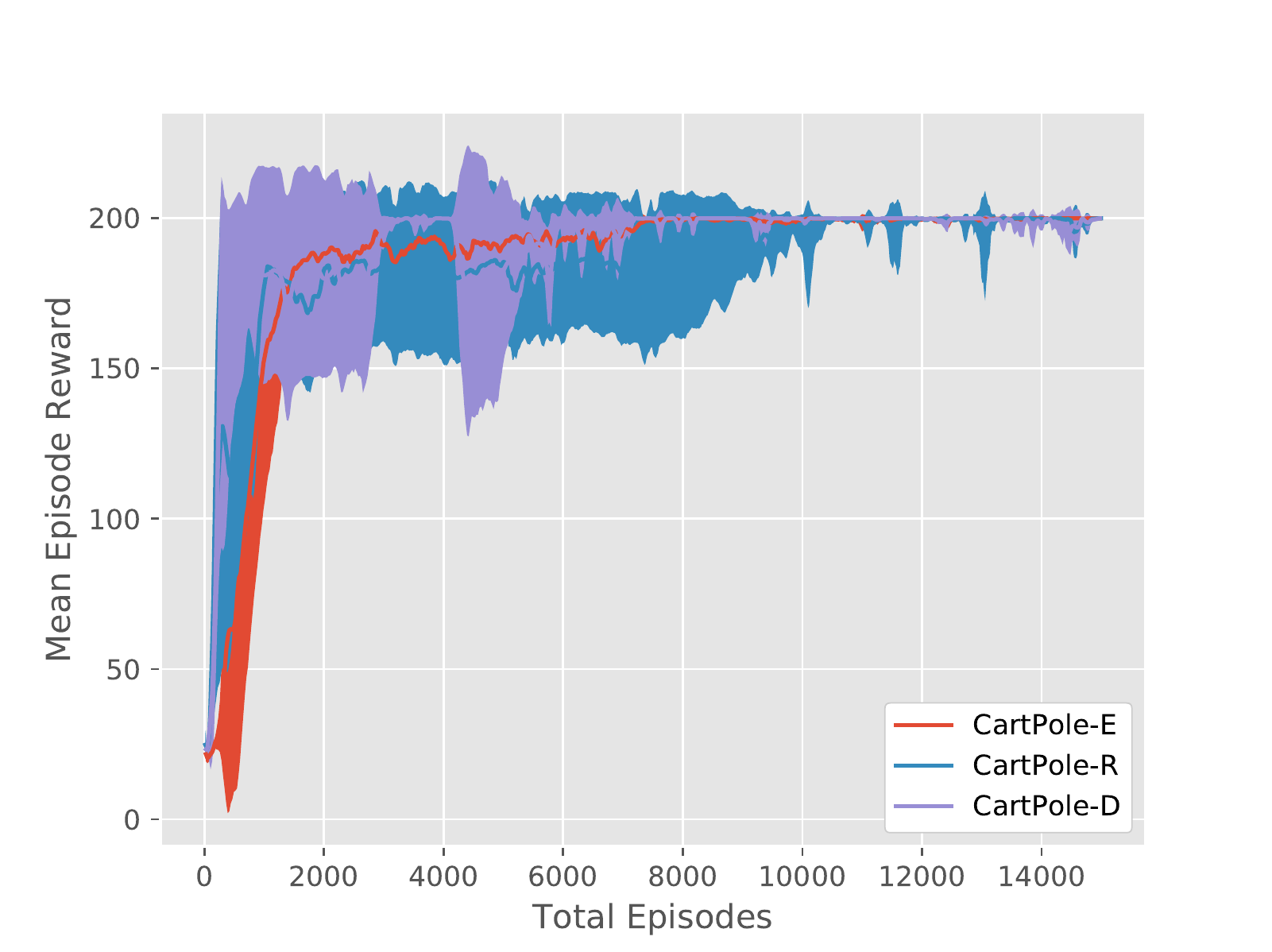}}
    \subfigure[PPO with RC architecture]{\includegraphics[width=\columnwidth]{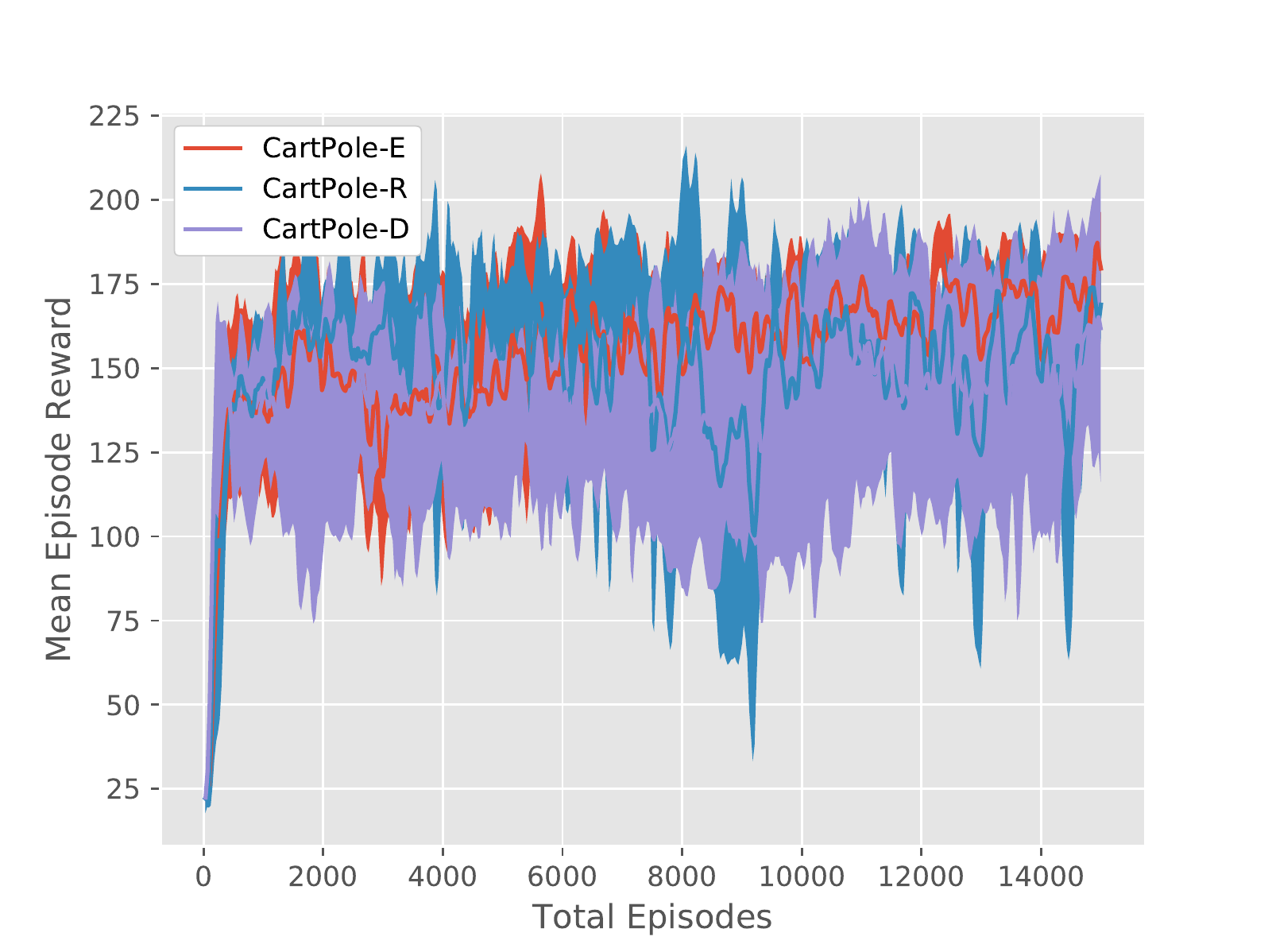}}
    \subfigure[EPOpt-PPO with FF architecture]{\includegraphics[width=\columnwidth]{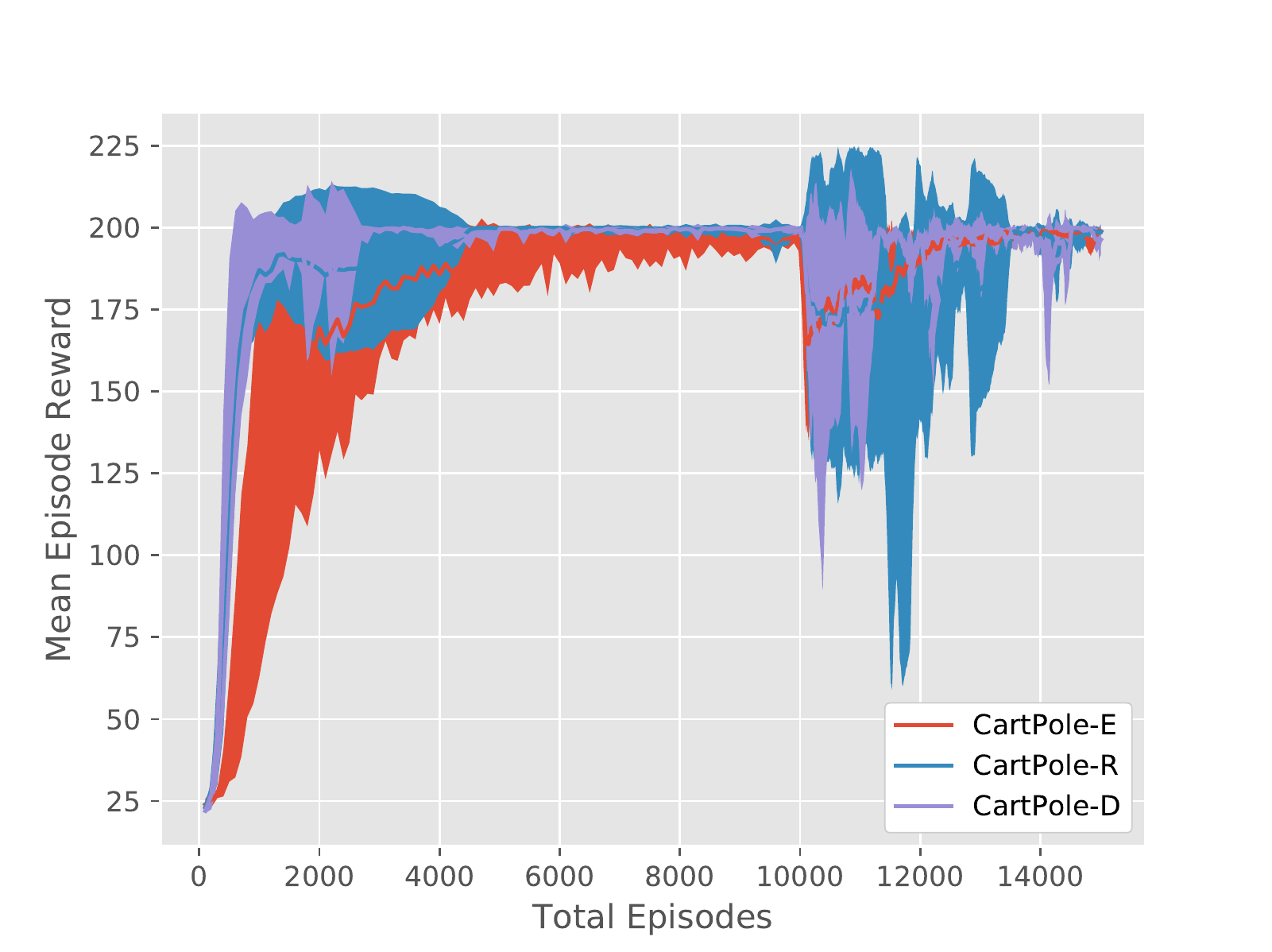}}
    \subfigure[EPOpt-PPO with RC architecture]{\includegraphics[width=\columnwidth]{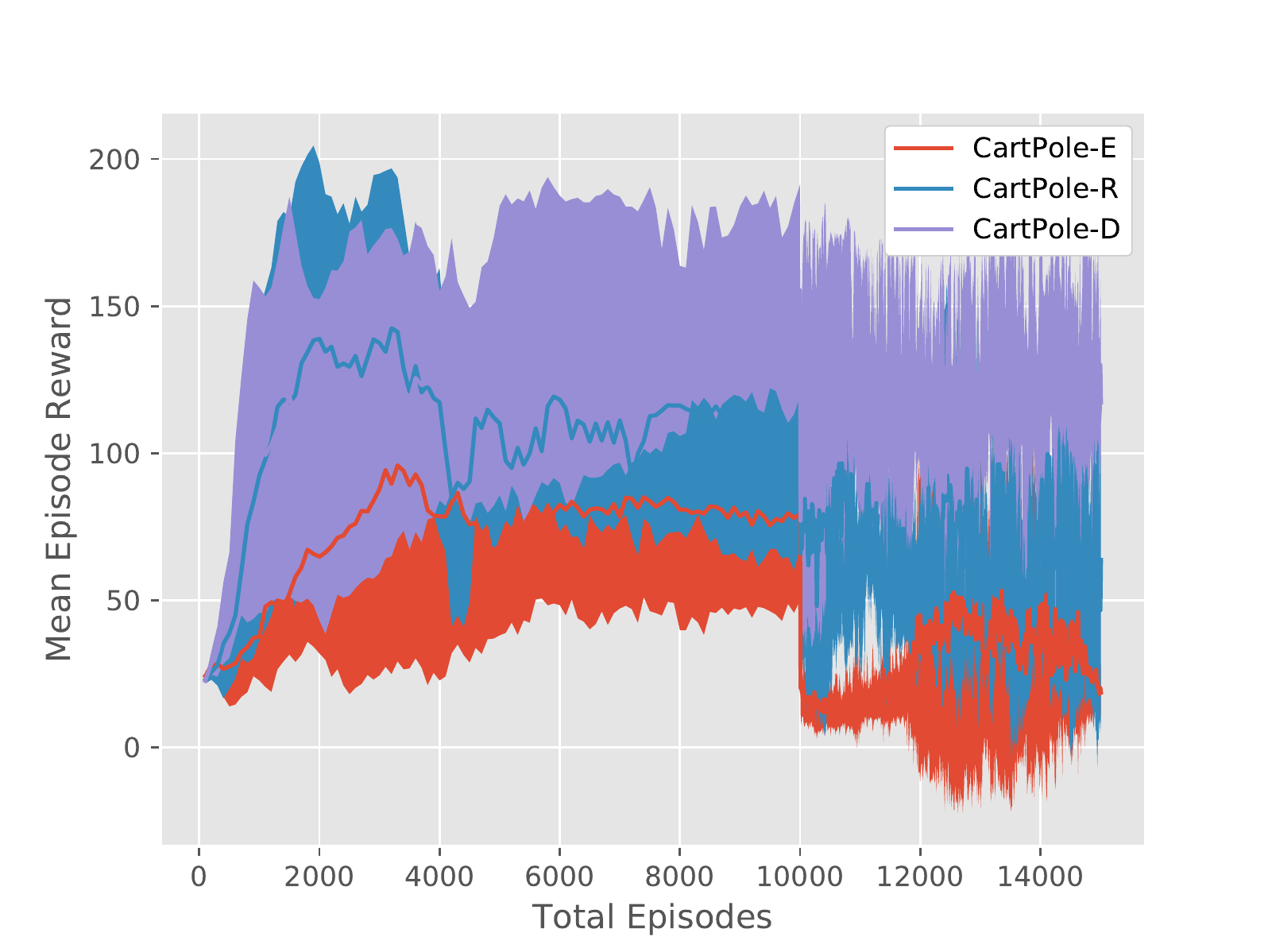}}
    \centering
    \subfigure[RL$^2$-PPO]{\includegraphics[width=\columnwidth]{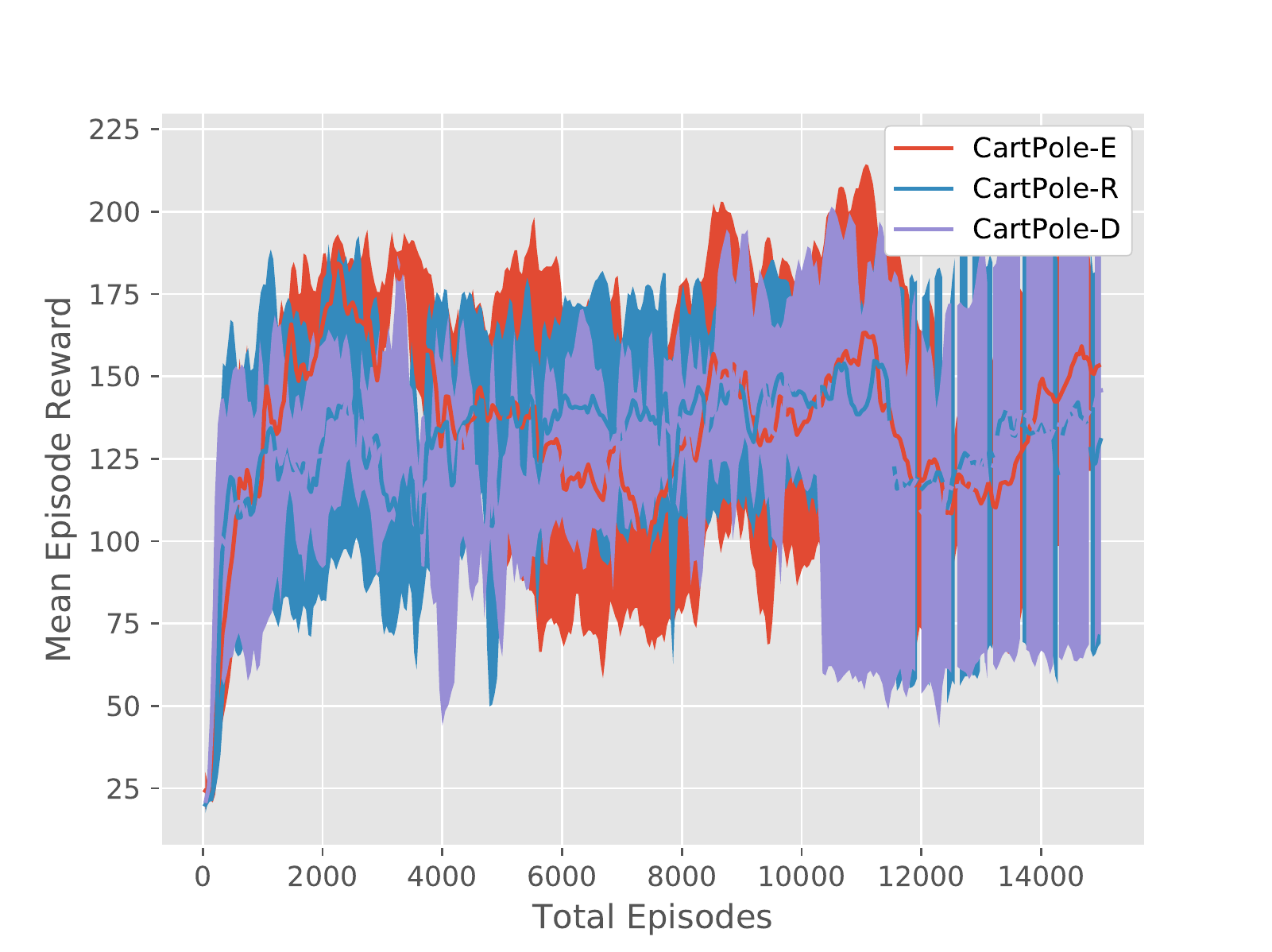}}

    \caption{Training curves for the PPO-based algorithms on CartPole, all three environment versions. Note that the decrease in mean episode reward at $10000$ episodes in the two EPOpt-PPO plots is due to the fact that it transitions from being computed using all generated episodes ($\epsilon=1$) to only the $10\%$ with lowest reward ($\epsilon=0.1$).}
    \label{fig:CartPoleTraining}
\end{figure*}

To investigate the effect of EPOpt and RL$^2$ and the different environment versions on training, we plotted the training curves for PPO, EPOpt-PPO, and RL$^2$-PPO on each version of each environment, averaged over the five experiment runs and showing error bands based on the standard deviation over the runs.
Training curves for all algorithms and environments are available at the following link: \href{https://drive.google.com/drive/folders/1H5aBv-Lex6WQzKI-a_LCgJUER-UQzKF4}{https://drive.google.com/drive/folders/1H5aBv-Lex6WQzKI-a\_LCgJUER-UQzKF4}.
We observe that in the majority of cases training appears to be stabilized by the increased randomness in the environments in R and E, including situations where successful policies are found.
This behavior is particularly apparent for CartPole, whose training curves are shown in Figure~\ref{fig:CartPoleTraining} and in which all algorithms above are able to find at least partial success.
We see that especially towards the end of the training period, the error bands for training on E are narrower than those for training on D or R\@. 
Except for EPOpt-PPO with the FF architecture, the error bands for training on D appear to be the widest. 
In particular, RL$^2$-PPO is very unstable when trained on D, possibly because the more expressive policy network overfits to the generated trajectories.

\section{Videos of trained agents}
\label{sec:videos}

The above link also contains videos of the trained agents of one run of the experiments for all environments and algorithms.
Using HalfCheetah as a case study, we describe some particularly interesting behavior we see.

\begin{figure*}[ht]
  \vskip 0.1in
  \begin{center}
  \includegraphics[width=1.6\columnwidth]{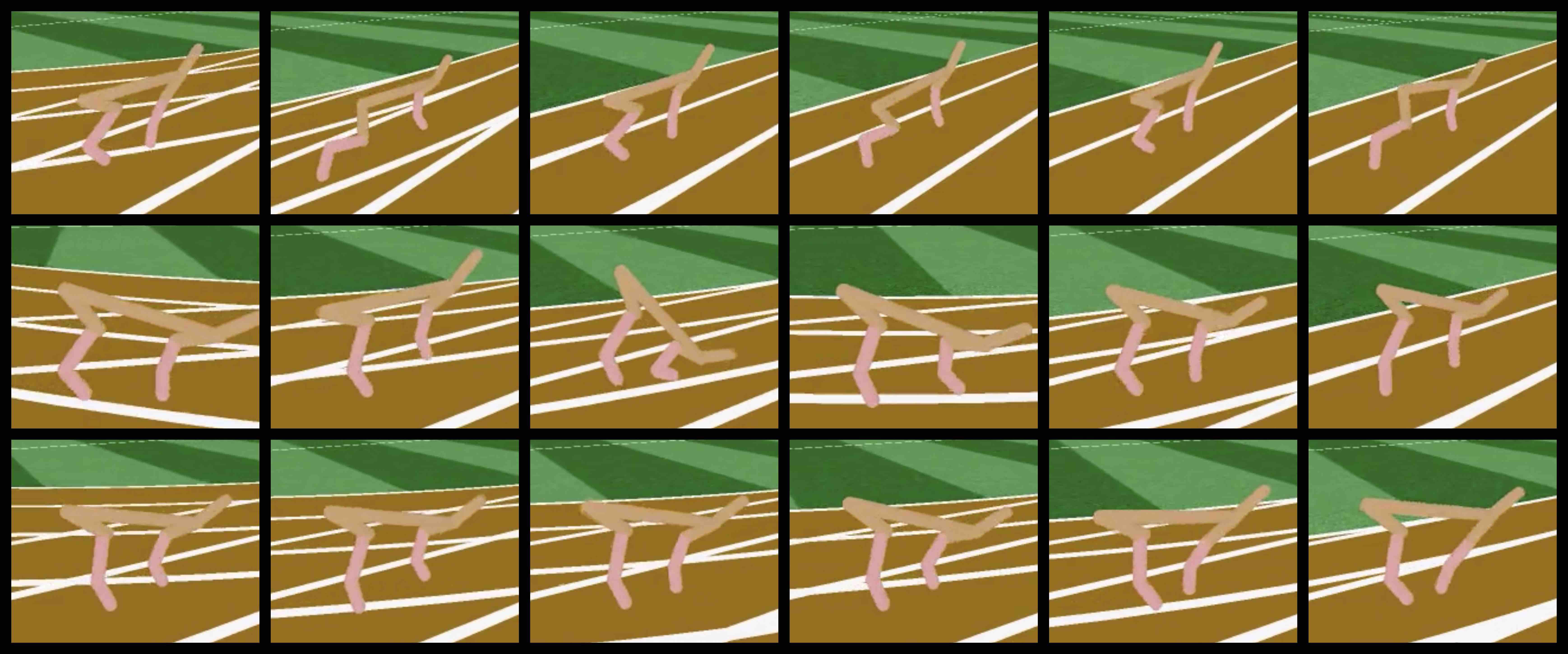}
  \caption{Video frames of agents trained with A2C on HalfCheetah, trained in the Deterministic (D), Random (R), and Extreme (E) settings (from top to bottom). All agents evaluated in the D setting.}
  \label{fig:a2cALL}
  \end{center}
  \vskip -0.1in
\end{figure*}

\begin{figure*}[ht]
  \vskip 0.1in
  \begin{center}
  \includegraphics[width=1.6\columnwidth]{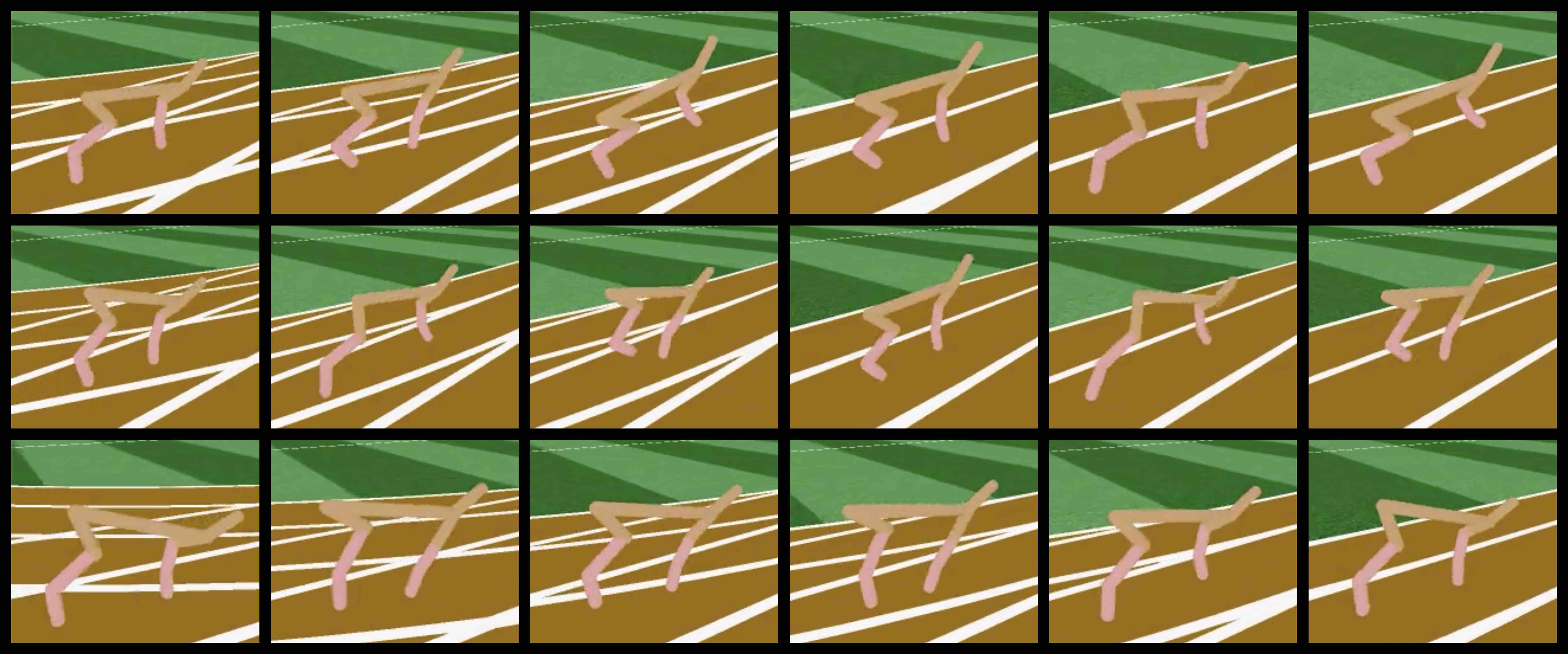}
  \caption{Video frames of agents trained with PPO on HalfCheetah, trained in the Deterministic (D), Random (R), and Extreme (E) settings (from top to bottom). All agents evaluated in the D setting.}
  \label{fig:ppoALL}
  \end{center}
  \vskip -0.1in
\end{figure*}

A trend we noticed across several algorithms were similar changes in the cheetah's gait that seem to be correlated with the difficulty of the environment. 
The cheetah's gait became forward-leaning when trained on the Random and Extreme environments, and remained relatively flat in the agents trained on the Deterministic environment (see figures \ref{fig:a2cALL} and \ref{fig:ppoALL}).
We hypothesize that the forward-leaning gait developed to counteract conditions in the R and E settings.
The agents with the forward-learning gait were able to recover from face planting (as seen in the second row of figure \ref{fig:a2cALL}), as well as maintain balance after violent leaps likely caused by settings with unexpectedly high power.
In addition to becoming increasingly forward-leaning, the agents' gait also tended to become stiffer in the more extreme settings, developing a much shorter, twitching stride.
Though it reduces the agents' speed, a shorter, stiffer stride appears to make the agent more resistant to adverse settings that would cause an agent with a longer stride to fall.
This example illustrates how training on a range of different environment configurations may encourage policies that are more robust to changes in system dynamics at test time.

\end{document}